\documentclass[a4paper, 10pt]{ieeeconf}      % Use this line for a4
\IEEEoverridecommandlockouts                              % This command is only
\usepackage{dblfloatfix}    % To enable figures at the bottom of page
\usepackage{microtype}
\usepackage{arydshln}
\usepackage{pbox}
\usepackage{epsfig}
\usepackage{graphicx}
\usepackage{amsmath}
\usepackage{amssymb}
\usepackage{arydshln}
\usepackage{algorithm}
\usepackage{algorithmicx}
\usepackage{algpseudocode}
\usepackage{mathtools}
\usepackage{dsfont}
\usepackage[table,xcdraw]{xcolor}
\usepackage{subcaption}
\usepackage{siunitx}
\usepackage{arydshln}
\usepackage{todonotes}
\usepackage{booktabs}
\usepackage[export]{adjustbox}
\usepackage[english]{babel}
\usepackage[%
    style=ieee,sorting=none,
    sortcites=true,doi=false,url=false,
    giveninits=true,minbibnames=2,maxbibnames=5,hyperref]{biblatex}
    
\algnewcommand{\LineComment}[1]{\State \# \textit{#1}}

\usepackage{array}
\newcolumntype{H}{>{\setbox0=\hbox\bgroup}c<{\egroup}@{}}

\newenvironment{itemize_tight}{
\begin{itemize}
	\setlength{\itemsep}{1pt}
	\setlength{\parskip}{1pt}
}{\end{itemize}}

\newcommand{\reff}[1]{Fig.~\ref{fig:#1}}

\usepackage[pdfusetitle,colorlinks]{hyperref}

\title{\LARGE \bf
Generating superpixels using deep image representations\\[.5em]
}

\author{Thomas Verelst \quad Matthew Blaschko \quad Maxim Berman \\
	Dept.\ ESAT, Center for Processing Speech and Images\\
 	KU Leuven, Belgium\\
{\tt\small \{thomas.verelst,matthew.blaschko,maxim.berman\}@esat.kuleuven.be} \\
}

\begin{document}

\maketitle
\thispagestyle{plain}
\pagestyle{plain}

%%%%%%%%%%%%%%%%%%%%%%%%%%%%%%%%%%%%%%%%%%%%%%%%%%%%%%%%%%%%%%%%%%%%%%%%%%%%%%%%
\begin{abstract}
Superpixel algorithms are a common pre-processing step for computer vision algorithms such as segmentation, object tracking and localization. 
Many superpixel methods only rely on colors features for segmentation, limiting performance in low-contrast regions and applicability to infrared or medical images where object boundaries have wide appearance variability. 
We study the inclusion of deep image features in the SLIC superpixel algorithm to exploit higher-level image representations. In addition, we devise a trainable superpixel algorithm, yielding an intermediate domain-specific image representation that can be applied to different tasks. A clustering-based superpixel algorithm is transformed into a pixel-wise classification task and superpixel training data is derived from semantic segmentation datasets. Our results demonstrate that this approach is able to improve superpixel quality consistently. 

\end{abstract}

%%%%%%%%%%%%%%%%%%%%%%%%%%%%%%%%%%%%%%%%%%%%%%%%%%%%%%%%%%%%%%%%%%%%%%%%%%%%%%%%%%%%%%%%%%%%
\section{Introduction}%\todo{I changed the page layout from letter to a4 so you might notice some unwanted changes. Also changed the title.}
Many deep learning based applications in computer vision operate on a grid of pixels and use convolutions trained end-to-end.
% However, vision tasks are often not sensitive to the value of each individual pixels. 
However, popular algorithms have successfully leveraged image segmentation to group pixels into superpixels, reducing the input dimensionality while preserving the semantic content needed to address the task at hand~\cite{fulkerson2009class}.
Superpixels are efficient image priors that tend to transfer across tasks and reduce the data needed to train models, which can be very beneficial for domain adaptation and weakly supervised settings, e.g. weakly supervised image segmentation \cite{kwak2017weakly}. Graph-based convolutional networks~\cite{graphbased} also allow applications of deep learning beyond grid-like inputs. Some works \cite{schuurmans2018efficient} explored the inclusion of superpixels in deep learning pipelines.
%In this work, we study the inclusion of superpixel priors in deep learning pipelines. 
%outside the scope of spatial grid convolutions.
% In this work, we study the inclusion of superpixels in deep learning pipelines. 

The hand-crafted design of superpixels algorithms limits our ability to tune image segmentations to a specific image domains, such as infrared, medical, of spatio-temporal data. Given the focus on efficiency, superpixels have often been designed to operate on color features only; image segmentations could however incorporate higher-level image representations. 
We consider extensions to a standard superpixel algorithms incorporating higher-level unsupervised or supervised image features. We also study paths to fine-tune a superpixel segmentation algorithm to a specific modality.
There has been few research on trainable superpixels. In parallel to our work, {Wei-Chih Tu\textit{ et al.}}~\cite{liulearning} have developed a trainable variant of graph-based superpixel algorithms using trainable superpixel affinities. Our approach is based on a clustering algorithm, which tends to be faster and more suited for real-time applications due to their iterative nature \cite{spix_eval}.

%%%%%%%%%%%%%%%%%%%%%%%%%%%%%%%%%%%%%%%%%%%%%%%%%%%%%%%%%%%%%
\section{SLIC algorithm}
Several comparisons indicate that the Simple Linear Iterative Clustering (SLIC)~\cite{slic} image segmentation algorithm offers both good speed and performance \cite{spix_eval}\cite{neubert2012superpixel}.
It uses a clustering approach similar to $k$-means, and usually operates on images in the CIELAB color space.
% is one of the most popular and recent algorithms, and uses a clustering-based approach similar to k-means. The input image is converted to the $CIELAB$ color space and SLIC starts by distributing $K$ cluster centers over the image. 
After initialization of the cluster centers along a grid, a two-step iterative process refines clusters until convergence. First, the pixels are assigned to the closest cluster center in a joint $5$-dimensional space of colors ($L$, $a$ and $b$) and spatial ($x$ and $y$) components. The weighted $L2$ distance includes a compactness parameter $\sigma$ to balance between colors and space. Second, the cluster centers are updated based on the pixel assignments. Finally, after convergence, a simple connected components algorithm enforces connectedness of the image segments.

%%%%%%%%%%%%%%%%%%%%%%%%%%%%%%%%%%%%%%%%%%%%%%%%%%%%%%%%%%%%

\section{Augmenting SLIC with deep representations}\label{sec:simpleintegration}
\subsection{Deep representations}
We experiment with SLIC beyond the original $Lab$ features. Deep representations capturing textures, gradients and edges in the image can be extracted from convolutional neural networks. Their structure is similar to multi-channel images, often having a lower resolution than the original image. Each channel represents an image feature. These features can be unsupervised, as in the case of scattering features~\cite{scattering} (\reff{scatfeat}), or trained for a particular vision task. Segmentation networks such as ENet \cite{paszke2016enet} have convolutional layers behaving like feature extractors. As we aim to integrate superpixels in a deep architecture, the features can be provided at no extra computational cost. 
Unsupervised scattering networks are similar to convolutional neural networks whose filters are fixed as wavelets. We use scattering networks with a receptive field of $4 \times 4$ for our experiments on $256 \times 256$ images, generating $M = 81$ features maps of size $64 \times 64$ per image channel.% We only use features on the muminance channel $L$. 

%We typically experiment with early layers of a CNN, which typically behave like smooth and universal filters. 

% We consider augmenting SLIC with supervised features extracted from early layers of a deep CNN. These layers typically behave like smooth and universal filters, detecting a specific pattern or texture. When using the superpixel priors in combination with a deep neural network, these features can be retrieved without additional cost.

% We also explore the usability of scattering networks~\cite{scattering}, which are similar to unsupervised CNNs with fixed wavelet filters. The filters are pre-defined and thus unsupervised, with few parameters. Spatial resolution is reduced as the number of features increases. 
%The scattering transform with depth 2 transforms a $256 \times 256 \times 1$ greyscale image into a $64 \times 64 \times 81$ image, where the features have a receptive field of 4 by 4.
% A few examples are shown in Figure \ref{fig:scatfeat}.

\begin{figure}
	\captionsetup{font=small}
	\captionsetup[sub]{font=footnotesize,labelfont={it,it}}
	\centering
	\begin{subfigure}[h]{0.24\linewidth}
		\includegraphics[trim={3.72cm 1.42cm 3.72cm 1.42cm},clip,width=\linewidth]{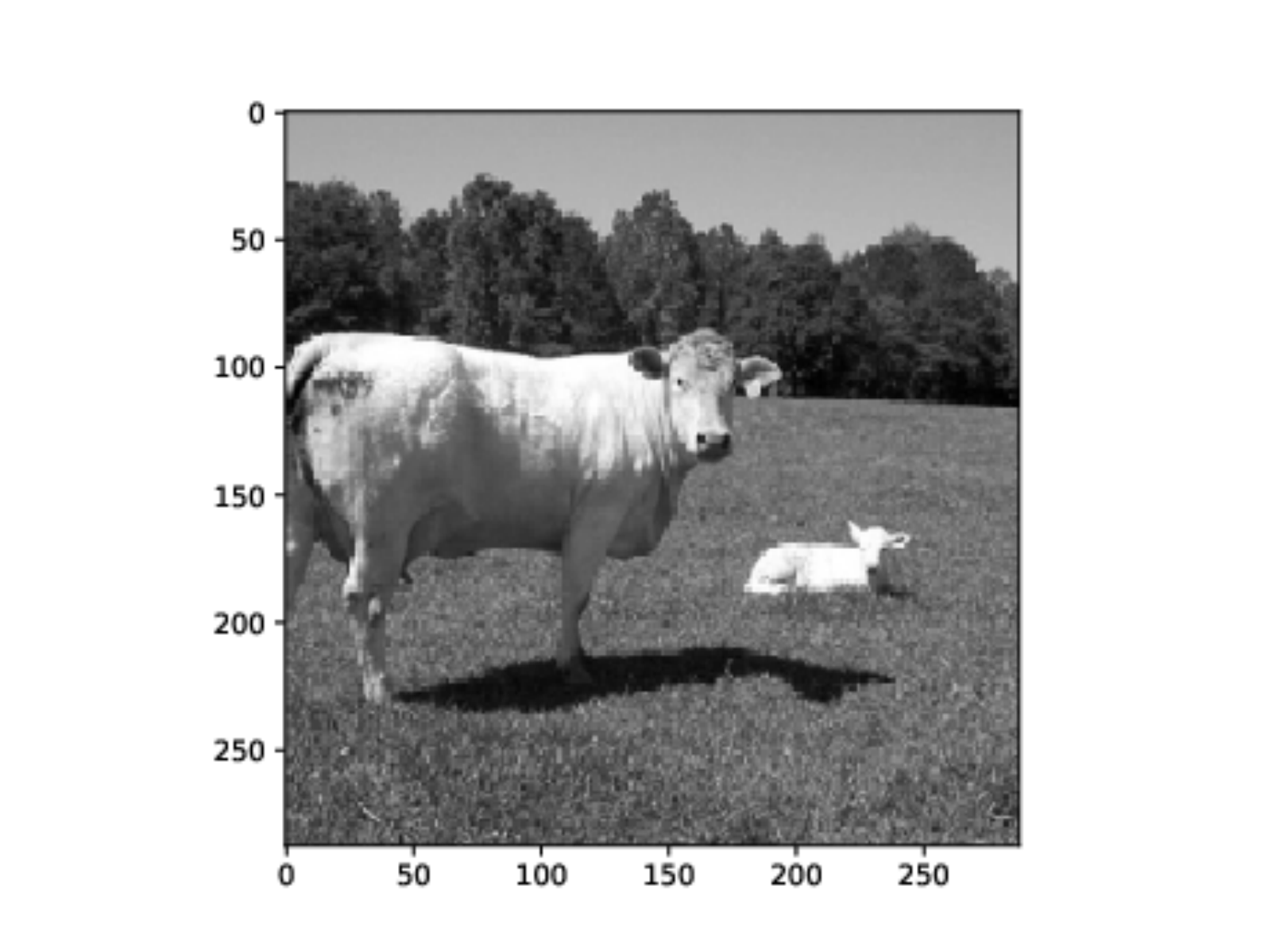}
	\end{subfigure}
	\hfill
	\begin{subfigure}[h]{0.24\linewidth}
		\includegraphics[trim={3.72cm 1.42cm 3.72cm 1.42cm},clip,width=\linewidth]{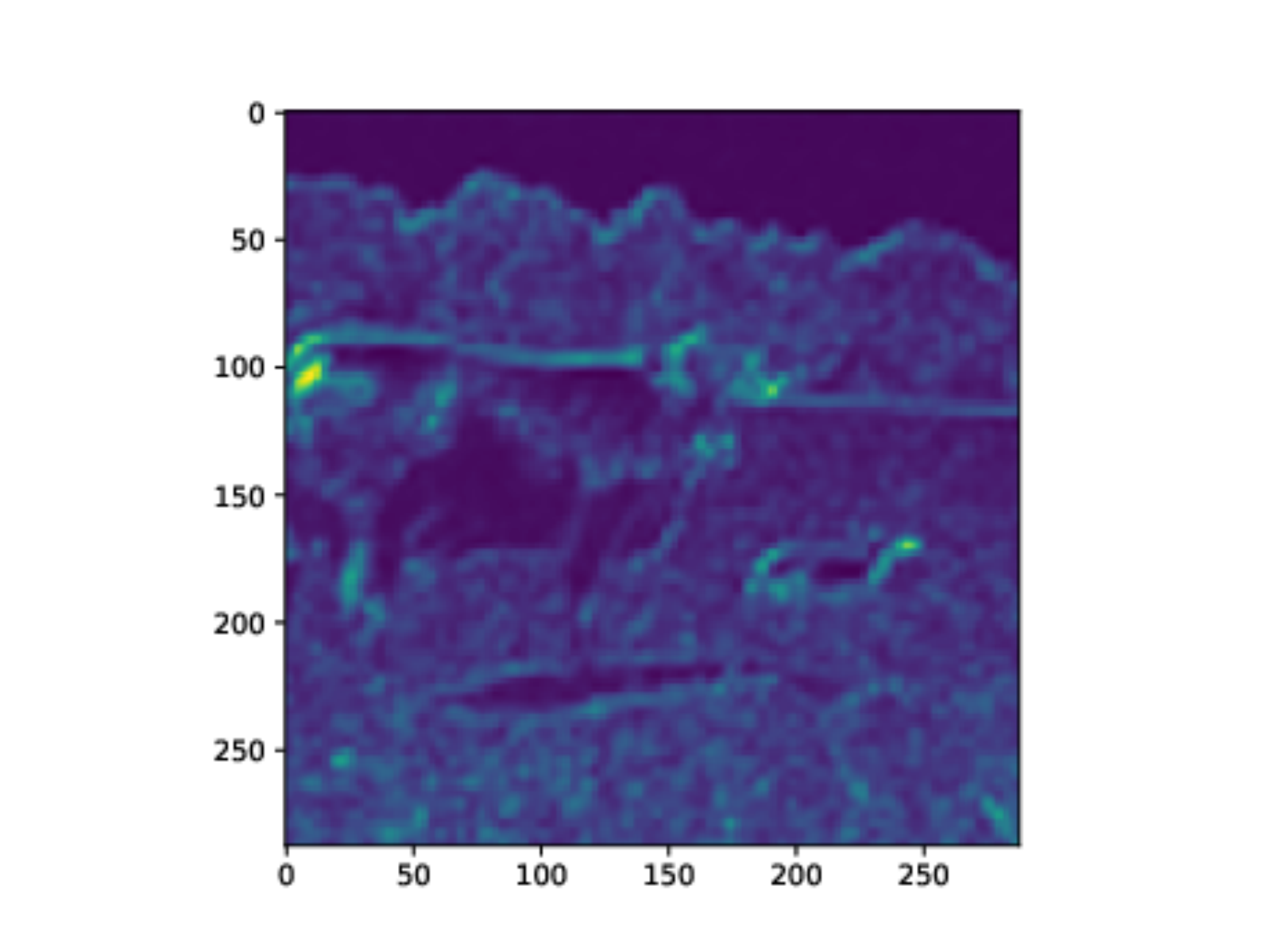}
	\end{subfigure}
	\hfill
	\begin{subfigure}[h]{0.24\linewidth}
		\includegraphics[trim={3.72cm 1.42cm 3.72cm 1.35cm},clip,width=\linewidth]{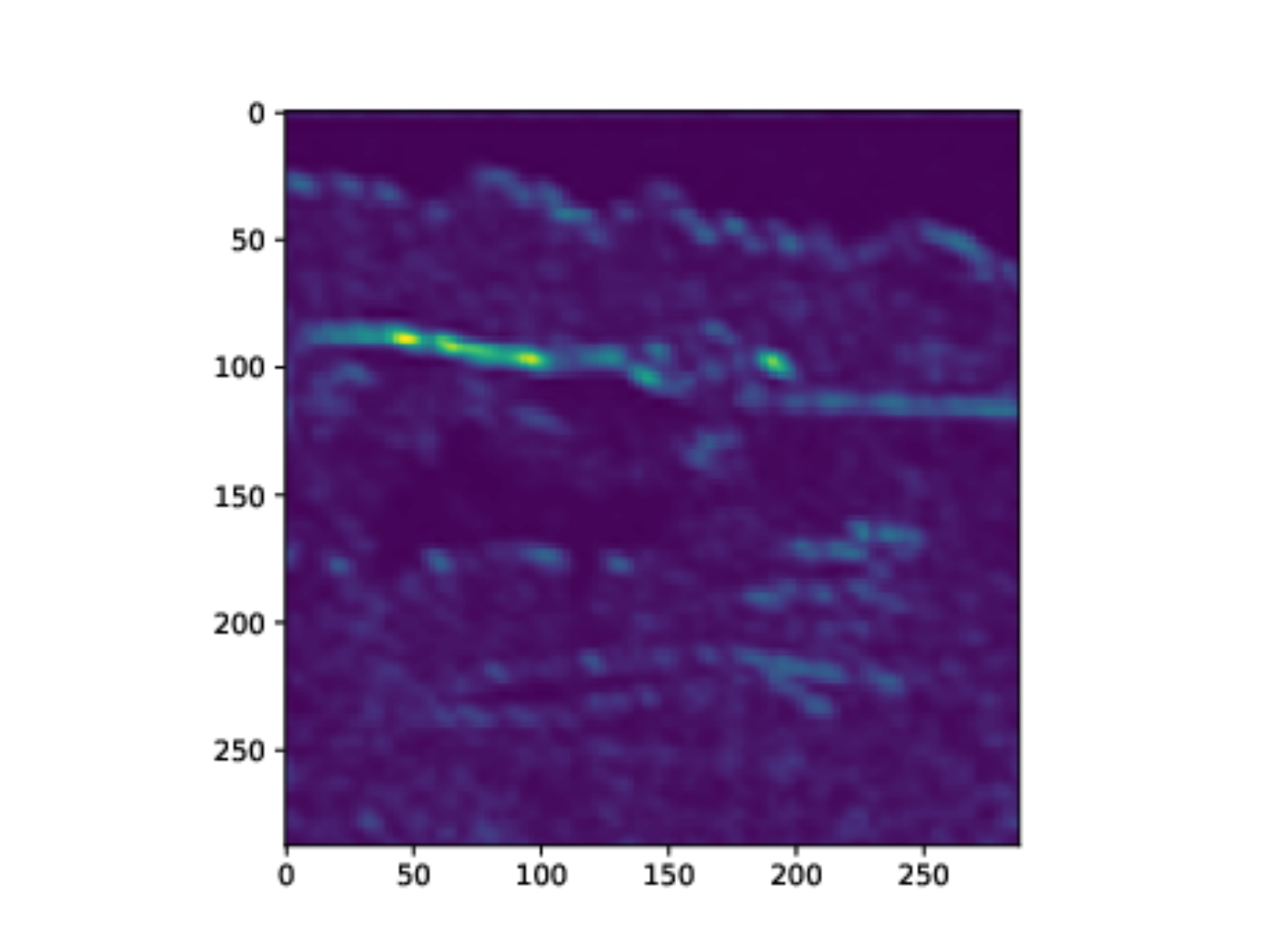}
	\end{subfigure}
    \hfill
	\begin{subfigure}[h]{0.24\linewidth}
		\includegraphics[trim={3.72cm 1.42cm 3.72cm 1.35cm},clip,width=\linewidth]{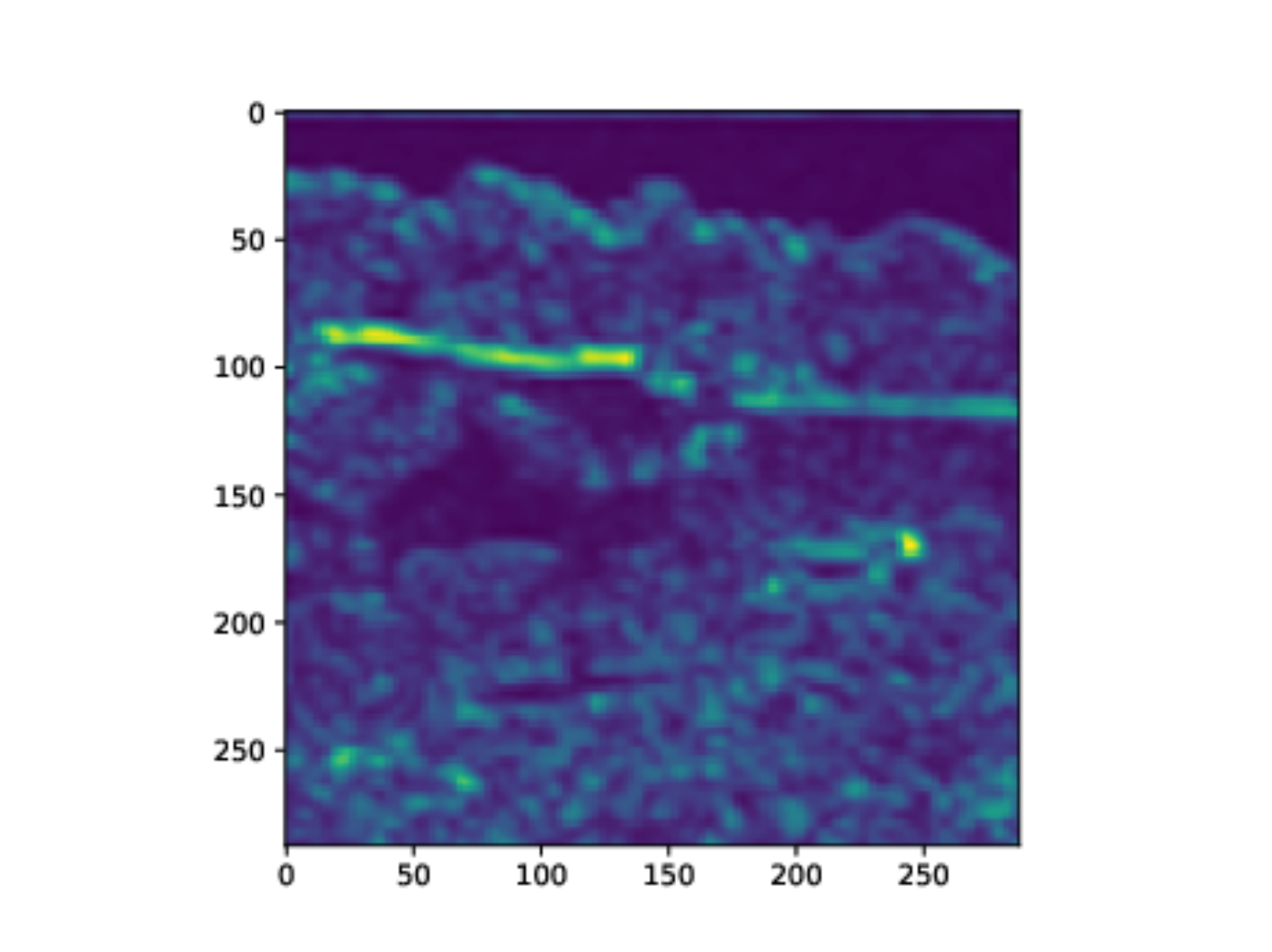}
	\end{subfigure}
	\caption{Input image and some scattering features}
	\label{fig:scatfeat}
\end{figure}

\subsection{Inclusion in SLIC}
For a particular pixel, we have image features $f_1,f_2, \ldots f_{M}$.
%retrieved by applying the scattering transform
%on the luminance component $L$ of the image. 
To incorporate the image features into SLIC, we augment the number of image channels. The scattering features are upscaled and concatenated with the input image (\reff{integrationarch}). The final image of size $W\times H\times (M+3)$ can be used in the SLIC algorithm, where the $Labxy$ clustering space now becomes a larger $L,a,b,f_1, \ldots, f_{M},x,y$ space.  The SLIC color distance is extended and individual feature maps are weighted with coefficients $\beta_1 ... \beta_M$. The distance in the color space between pixel $i$ and cluster $k$ is then defined as
\begin{equation}
\begin{aligned}
d_c^{2} =& \alpha_1(L_k- L_i)^2 + \alpha_2(a_k - a_i)^2 + \alpha_3(b_k - b_i)^2 \\
&+ \sum_{m=1}^M \beta_m(f_{m, k} - f_{m, i})^2\ .
%\\ \beta_0(f_{0,k} - f_{0,i})^2 + \beta_1(f_{1,k} - f_{1,i})^2 + .. + \beta_{M-1}(f_{M-1,k} - f_{M-1,p})^2. 
	\label{formula:slicweighteddsitance}
\end{aligned}
\end{equation}

\begin{figure}[t]
\captionsetup{font=small}
	\captionsetup[sub]{font=footnotesize,labelfont={it,it}}
\begin{center}
\includegraphics[width=1\linewidth]{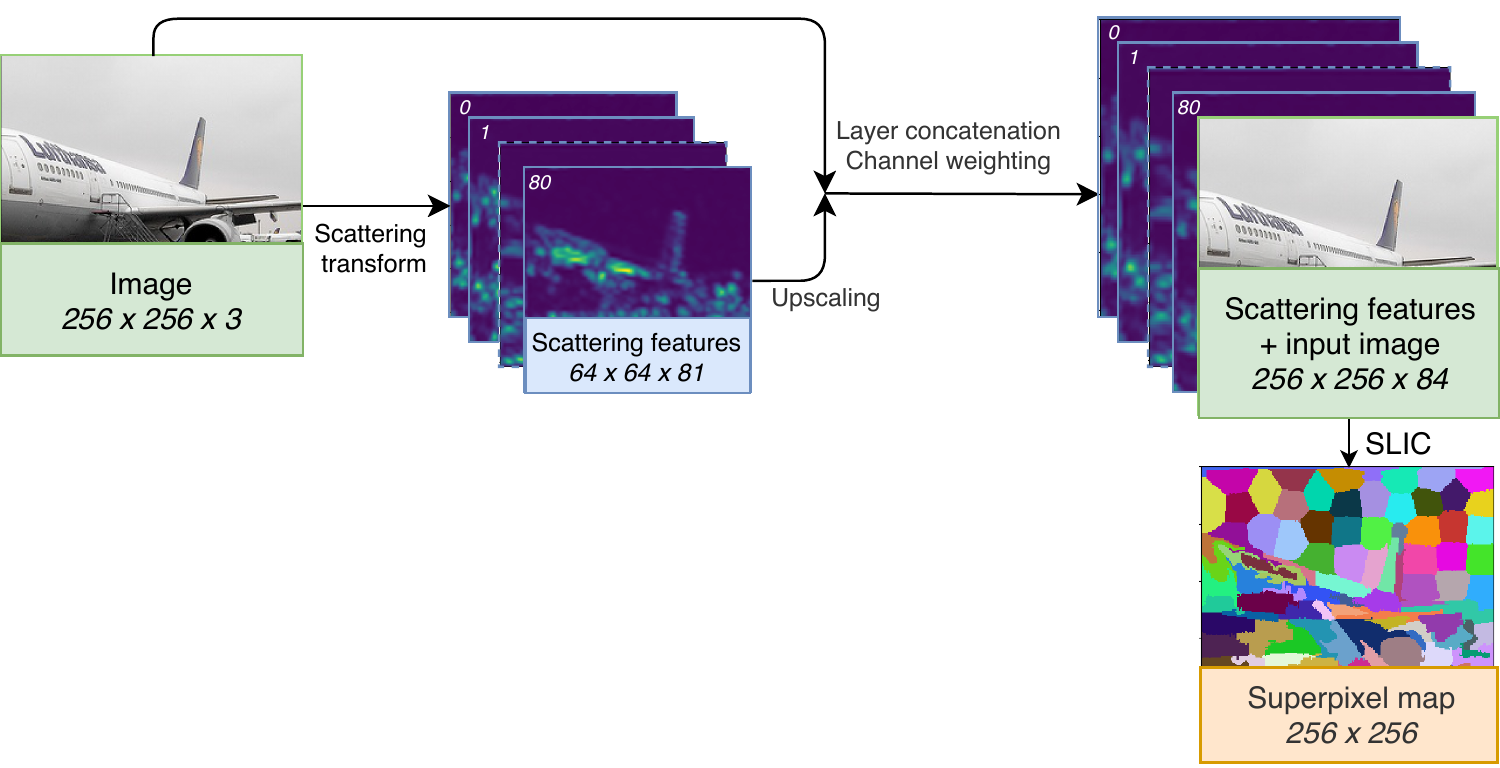}
\end{center}
   \caption{The feature maps are upscaled and concatenated with the original image. A SLIC algorithm with extended color distance measure is applied on the multi-channel image. }
\label{fig:integrationarch}
\end{figure}

Our first experiments investigate the impact of scattering features by manually tuning the inclusion of scattering features on the lightness component $L$ only. We define binary weights $\beta_m$ based on the visual appearance of the features. Layers originating from strong edge detectors are left out since they are of no use in clustering. We also varied the relative importance of the extra features compared to the color components and selected the best-scoring approach (out of 10 different ones) for evaluation (Section~\ref{sec:manualevaluation}).

%%%%%%%%%%%%%%%%%%%%%%%%%%%%%%%%%%%%%%%%%%%%%%%%%%%%%%%%%%ùùùù

\section{Trainable superpixel algorithm}
Manual selection and weighting of features in the distance measure is a tedious process, requiring visual examination of features and an exhaustive search for optimal weights. In addition, the distance measure (\ref{formula:slicweighteddsitance}) might not have enough flexibility to integrate those features properly. We research a trainable superpixel algorithm incorporating a neural network that can tune superpixels to a certain image set.

\subsection{Clustering as a classification problem }
% We use a more rigid framework generating valid superpixel structures without post-processing. 
The SLIC superpixel algorithm uses a top-down approach: the algorithm iterates over all cluster centers and calculates a distance measure to all pixels in the $2S \times 2S$ neighborhood around the cluster center. An equivalent bottom-up approach would be to iterate over all pixels and calculate a distance measure between the pixel and all the clusters in the $2S \times 2S$ region around the pixel. The pixel is then assigned to the cluster being the closest in the $5D$ clustering space of SLIC with $Labxy$ components. 

This is in fact a classification problem: assign each pixel to one of the clusters in the spatial neighborhood. While SLIC solves this classification problem using a distance measure, we rather avoid to train a regression because distances improving superpixel performance are hard to define. We propose to use a neural classifier for the assignment task: it considers a fixed amount of spatially closest clusters in the neighborhood and assigns the pixel to one those depending on their features.

\subsection{Bottom-up trainable superpixel algorithm}
The algorithm (Algorithm~\ref{alg:pixelwise_training}) works in a similar way to SLIC. Clusters are first initialized on a grid. Then, clusters are formed using a two-step iterative procedure: the first step assigns each pixel to one of the $Q$ spatially closest clusters, using classification based on the features of these clusters and the pixel of interest. $Q$ is a parameter: higher means more flexibility at the cost of more computations.  Afterwards, the features and position of the newly formed clusters are calculated by averaging the features and positions of the pixels assigned to those clusters. This iterative procedure is done for a fixed amount of iterations. Finally, a connected components algorithm is used to transform clusters into proper superpixels. 

A sequential implementation as described here would be slow: the large amount of individual network evaluations limits the performance. We implemented a version that generates large batches and evaluates these on a GPU. The algorithm can also easily be parallelized because every pixel is processed independently. 

%\algsetup{linenosize=\small}
\begin{algorithm}[tb]
	%\SetAlFnt{\small}
    %\SetAlCawww
    \small
	\begin{algorithmic}
		\LineComment{Initialization}
		\State Initialize cluster centers \( C_k = [l_k, a_k, b_k, x_k, u_k]^T \) by sampling pixels at regular grid steps $S$.
		%\State Move cluster centers to lowest gradient position in a $3 \times 3$ neighborhood.\\
		
		\State label $l(i) \gets -1$ for each pixel $i$		
		
		\Repeat 
		\LineComment{Clustering iteration}
		\ForAll{pixels $p_i$}
		
				%\LineComment{returns a vector of size $C$, containing the indexes of the $C$ cluster centers} \LineComment{spatially closest to pixel $p$}
		
		%\LineComment{Find closest clusters}
		\State closest\_idx $\in \mathbb{R}^{1\times Q} \gets$ 
		{Q cluster centers nearest to $p_i$}%\\
		
		%\LineComment{Build input vector}
		\State input $\gets$ {features of pixel $p_i$ and Q nearest clusters }%\\\in \mathbb{R}^{(M + 3) + (M+3+1) \times C}\ 

		%\LineComment{Run neural network}
		%\LineComment{The output is a $1 \times C$ vector}
		\State output $\in \mathbb{R}^{1\times Q}\ \gets$ \textbf{network}(input)%\\
		
		%\LineComment{Assignment}
		%\LineComment{Get the index of the best cluster, based on classifier output}
		\State best\_cluster\_id $\gets$ closest\_idx[\textbf{argmax}(output)]
		\State $l(i) \gets$ best\_cluster\_id%\\

		\EndFor
		\State Compute new cluster centers and features based on $l$
		\Until{number of iterations reached}
		
	\caption{Bottom-up trainable superpixel algorithm}
	\label{alg:pixelwise_training}
	\end{algorithmic}
\end{algorithm}

\subsection{Neural network architecture}
The input vector for the classification of a single pixel consists of several parts:
\begin{itemize_tight}
	\item $M$ pixel features, for example the pixel color and other features extracted using deep representations. %In SLIC, the values at the pixel are not used
	\item $Q$ spatial distances to the $Q$ closest clusters. In order to have a single neural network for multiple superpixel sizes and compactness parameters, the distance is normalized: $D_{q} = {\sigma * distance_{q}}/{step\_size} $, 
	 with $distance_{q}$ the pixel distance between pixel and cluster center $k$.% The compactness parameter $\sigma$ can change the weight of the distance.
	\item $Q \times M$ feature differences between the input pixels and cluster centers.  
\end{itemize_tight}

The network outputs a vector of size $Q$, where each element $q = 1...Q$ denotes the probability of the pixel belonging to cluster with index $q$. 
We aim for a small network and look at the problem as a typical classification problem. A fully connected network would not exploit the similarity between different parts of the input vector. An efficient architecture is made up of three parts: normalization, dimensionality reduction and classification (\reff{inputvector_asdr}). The Dimensionality Reducer for Pixels (DRP) modules transforms the pixel features to a smaller space, while the Dimensionality Reducer for Clusters (DRC) is applied on the pixel-cluster differences. Weights are shared between similarly-named modules to reduce the number of trainable parameters. The final fully connected network (FC) does the actual classification. 

\begin{figure}[tb]
	\captionsetup{font=small}
	\captionsetup[sub]{font=footnotesize,labelfont={it,it}}
	\centering
	\includegraphics[width=\linewidth]{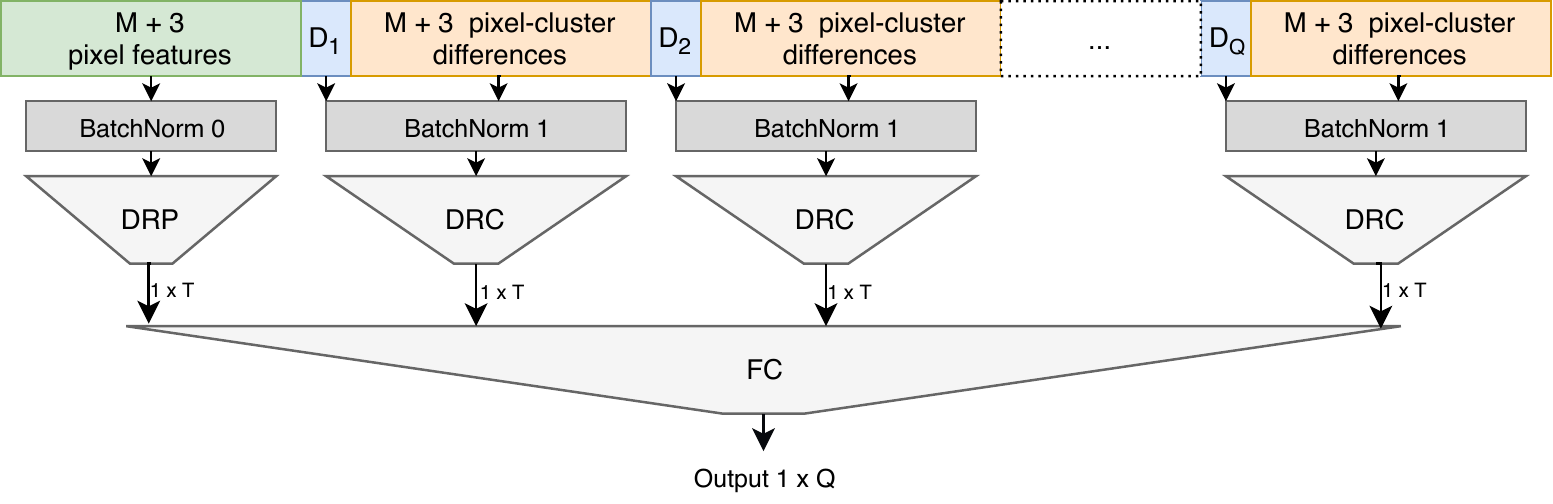}
	\caption{Classifier with dimensionality reduction modules. The weights
of the Dimensionality Reducer for Clusters (DRC) modules are shared to reduce the
amount of trainable weights.}
	\label{fig:inputvector_asdr}
\end{figure}
\begin{figure}[tb]
	\captionsetup{font=small}
	\captionsetup[sub]{font=footnotesize,labelfont={it,it}}
	\centering
	\includegraphics[width=0.85\linewidth]{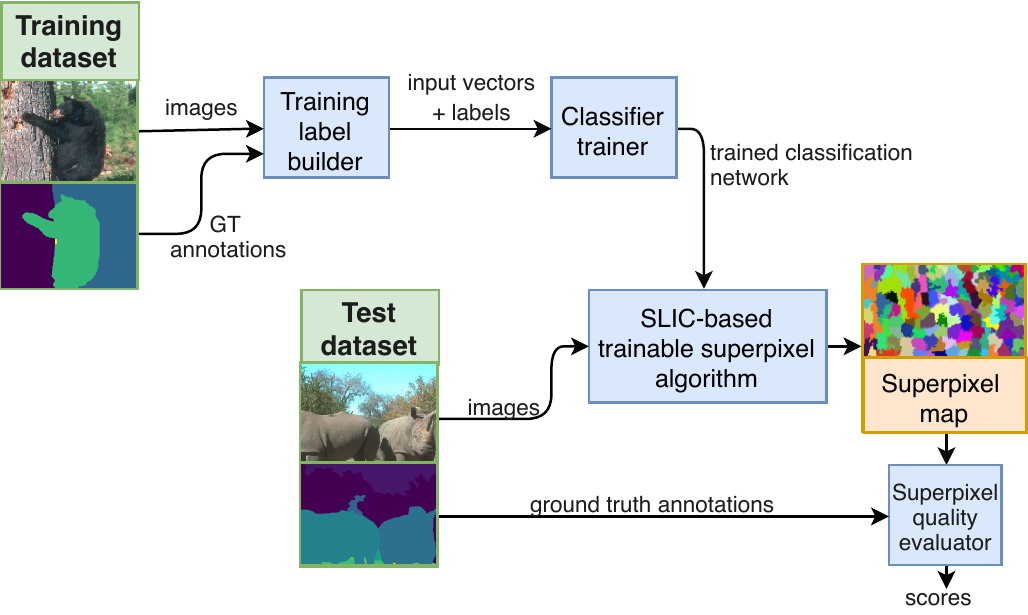}
	\caption{Diagram of the training and evaluation procedure. Training data for the classifier is derived from a semantic segmentation dataset. The input vectors and labels are used to train the classifier. Afterwards, the trained classifier is integrated in the trainable superpixel algorithm and results are evaluated on the test dataset.}
	\label{fig:evaluationlayoutdiagram}
\end{figure}

\section{Generating training labels}
The classifier requires training labels, indicating which cluster the pixel should be assigned to according to ground truth. Since no database with superpixel annotations exist, we derive a label set from semantic segmentation databases such as Cityscapes \cite{cityscapes} and BSDS \cite{bsds500} (\reff{evaluationlayoutdiagram}). % We can use improved labels based on SLIC, or more freedom with weakly supervised labelings. 

\subsection{SLIC-based labels}
We use the SLIC distance measure as a starting point to produce labels. SLIC replication requires to calculate the SLIC distance measure to the $Q$ closest clusters of the classifier and pick the closest cluster according to this measure. The pixel label is then set to this cluster. %This will yield slightly different results than the original SLIC algorithm because the amount of considered clusters is set to $Q$, while SLIC considers all clusters in the $2S \times 2S$ region around a pixel. 
Replicating SLIC would not force the classifier to include the features extracted from deep representations in its decision process. To improve superpixels beyond SLIC, we use ground truth annotations for semantic segmentation to correct wrong labels, where the pixel would be assigned to a cluster in a different ground truth segment. SLIC makes these mistakes when regions have approximately the same colors, but the classifier can use deep representations to discriminate between the two regions. When generating a label for a cluster, we only consider assignment to clusters lying mainly in the same ground truth segment as the pixel being classified. 

Ground truth segmentations are typically much larger than superpixels and the amount of pixels being corrected by the ground truth segmentation is small. The classifier thus primarily replicates SLIC and ignores the corrected labels. A multi-label loss could take into account that multiple clusters are good candidates, but we couldn't achieve satisfactory results using this approach. We solve the problem by using principles of hard-example mining: the set of labels is carefully chosen to improve the training process.

\paragraph{Hard-example mining on SLIC mistakes}
We try to train the classifier by only retaining labels that were corrected by the ground truth annotations. Our experiments indicate that this is too strict and degrades superpixel performance.

\paragraph{Hard-example mining at segmentation edges}
A less strict method would be to only consider pixels near ground truth edges. Labels in the middle of the ground truth segments have a lot of ambiguity: we cannot be sure whether the assigned cluster is really in the same part of the object. Labels at the edges have more discriminative power. We call these \textit{unambiguous labels}.
Our implementation does not exactly select pixels near the edge; it is easier to count the amount of different ground truth segments of the $Q$ closest clusters. % considered when classifying a pixel. 
Thus, we restrict the training set to pixels that have candidate clusters in at least a chosen amount of different ground truth segments.
% (\reff{unambiguouslabels}).  Interestingly, our experiments indicate that a higher value for $X$ produces more compact clusters (\reff{unambiguousoutput}). The reduced amount of ambiguity increases the importance of the spatial component: the network learns that two pixels next to each other might have very different features, while having very similar spatial distances the spatially closest clusters.

\subsection{Weakly supervised labeling}
Using the SLIC distance measure to generate pixel labels offers a good starting point but might also restrict the adaptability of the classification network. One could label a pixel to a random cluster in the same segment. This obviously generates very noisy superpixels. Picking the closest cluster in the same segment has the opposite problem: the spatial component is emphasized too much. Again, we leverage the principles of hard-example mining to build a better training set. We limit the training set to pixels having candidate clusters in at least $X$ segments, with an optimal $X$ to be determined experimentally (\reff{unambiguouslabels}). Interestingly, our experiments indicate that a higher value for $X$ produces more compact clusters (\reff{unambiguousoutput}). The reduced amount of ambiguity increases the importance of the spatial component: the network learns that two pixels next to each other might have very different features, while having very similar spatial distances to the spatially closest clusters.

\begin{figure}[H]
    \captionsetup{font=small}
	\captionsetup[sub]{font=footnotesize,labelfont={it,it}}
	\captionsetup[subfigure]{labelformat=empty}
	\centering
	\begin{subfigure}[h]{0.32\linewidth}
		\includegraphics[trim={2.72cm 1.42cm 2.72cm 1.42cm},clip,width=\linewidth]{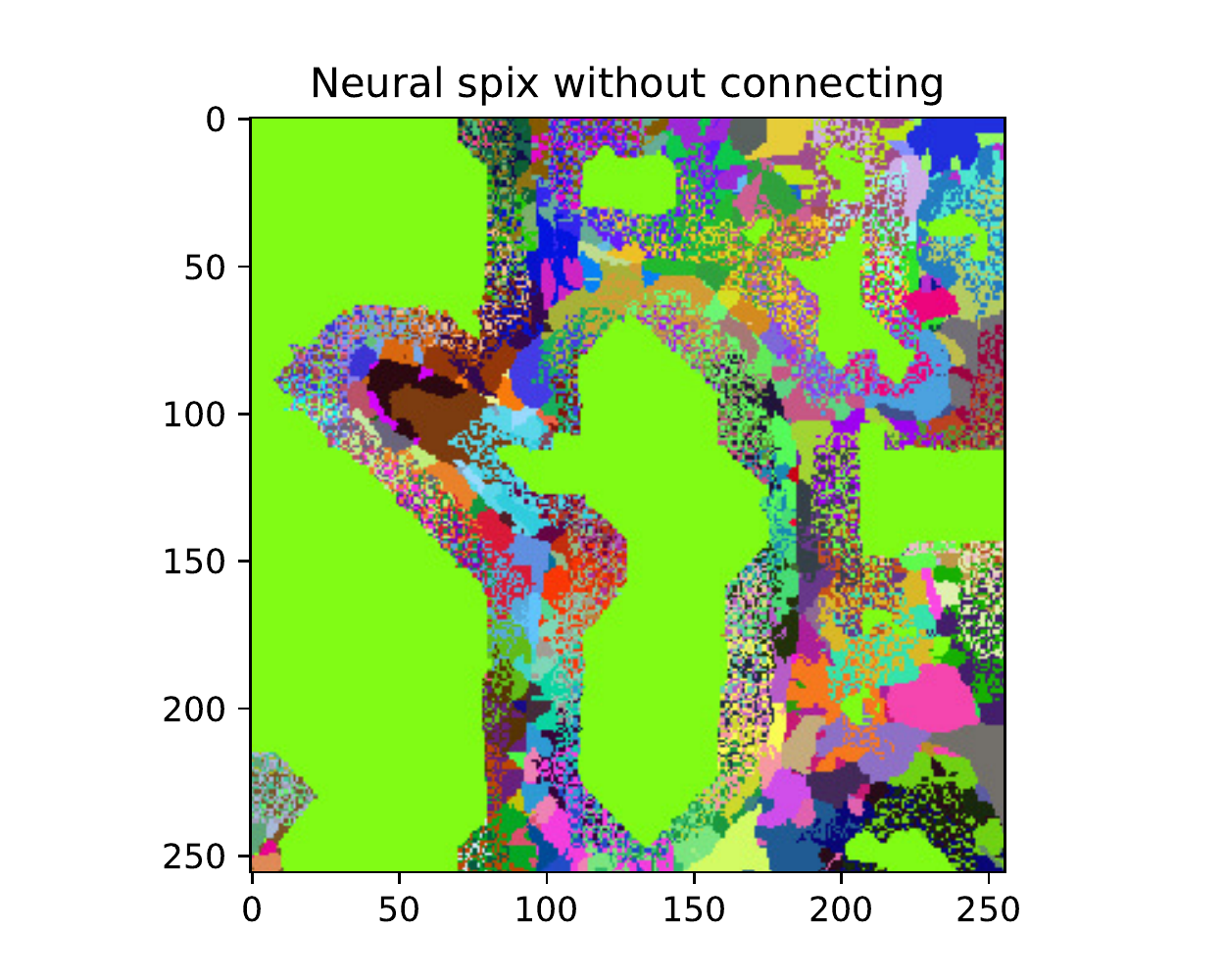}
		\subcaption{At least 2 different distances}
	\end{subfigure}
	\hfill
	\begin{subfigure}[h]{0.32\linewidth}
		\includegraphics[trim={2.72cm 1.42cm 2.72cm 1.42cm},clip,width=\linewidth]{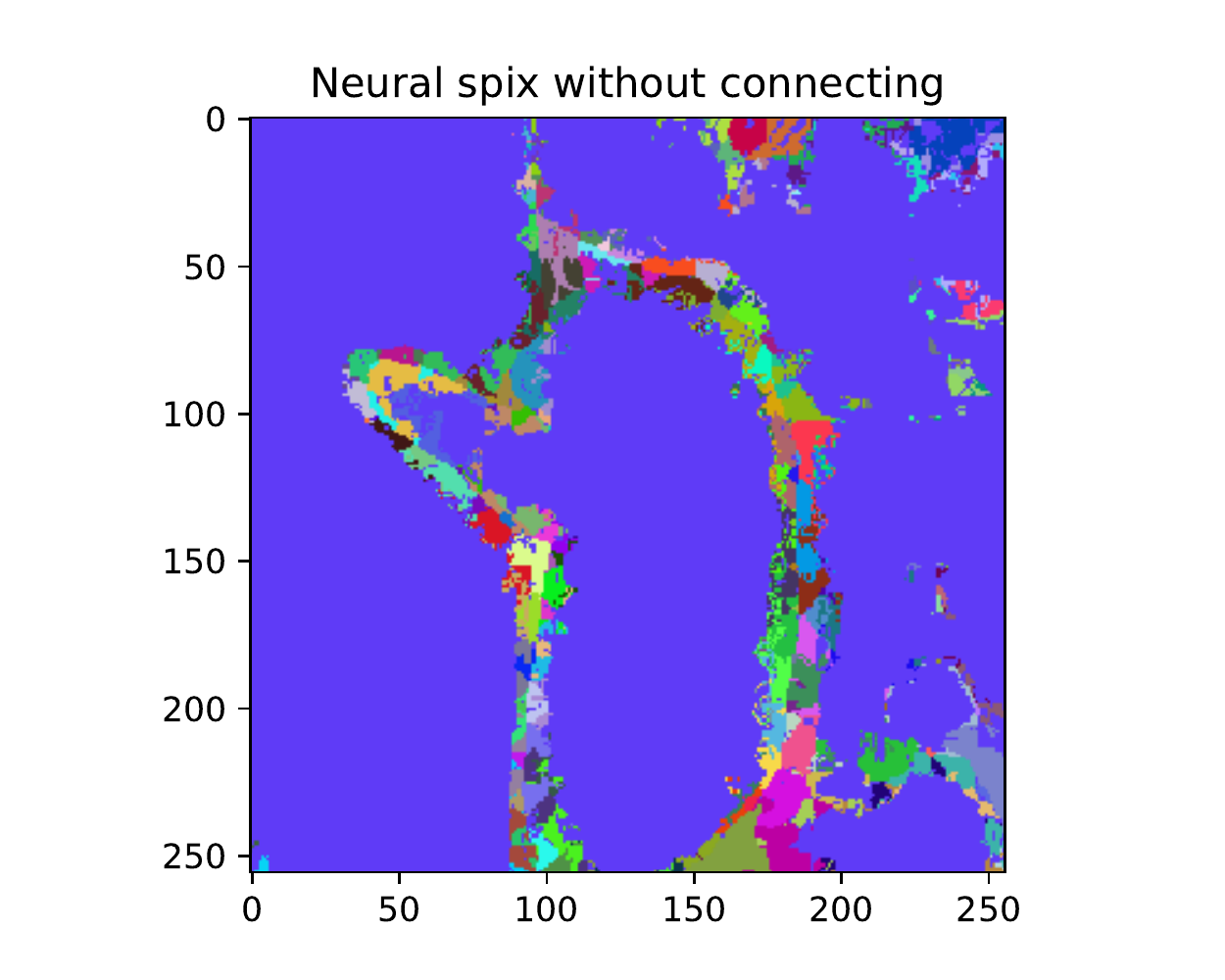}
		\subcaption{At least 4 different distances}
	\end{subfigure}
	\hfill
	\begin{subfigure}[h]{0.32\linewidth}
		\includegraphics[trim={2.72cm 1.42cm 2.72cm 1.42cm},clip,width=\linewidth]{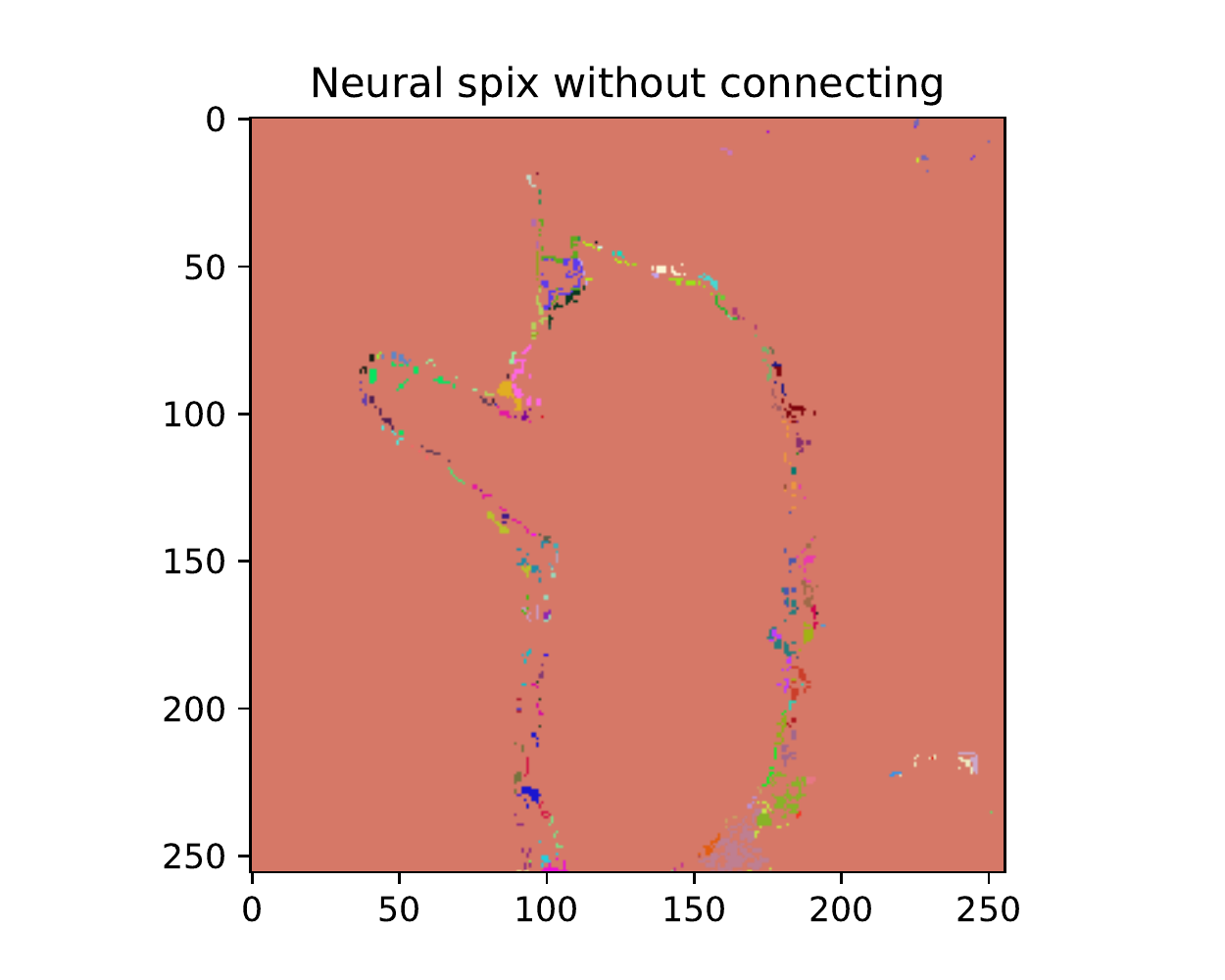}
		\subcaption{At least 6 different distances}
	\end{subfigure}
	\caption{Hard-example mining: only labels produced by clusters in minimum $X$ different ground truth segments are used for training.}
	\label{fig:unambiguouslabels}
\end{figure}
\begin{figure}[H]
	\captionsetup{font=small}
    \captionsetup[sub]{font=footnotesize,labelfont={it,it}}
	\captionsetup[subfigure]{labelformat=empty}
	\centering
	\begin{subfigure}[h]{0.32\linewidth}
		\includegraphics[trim={2.72cm 1.42cm 2.72cm 1.42cm},clip,width=\linewidth]{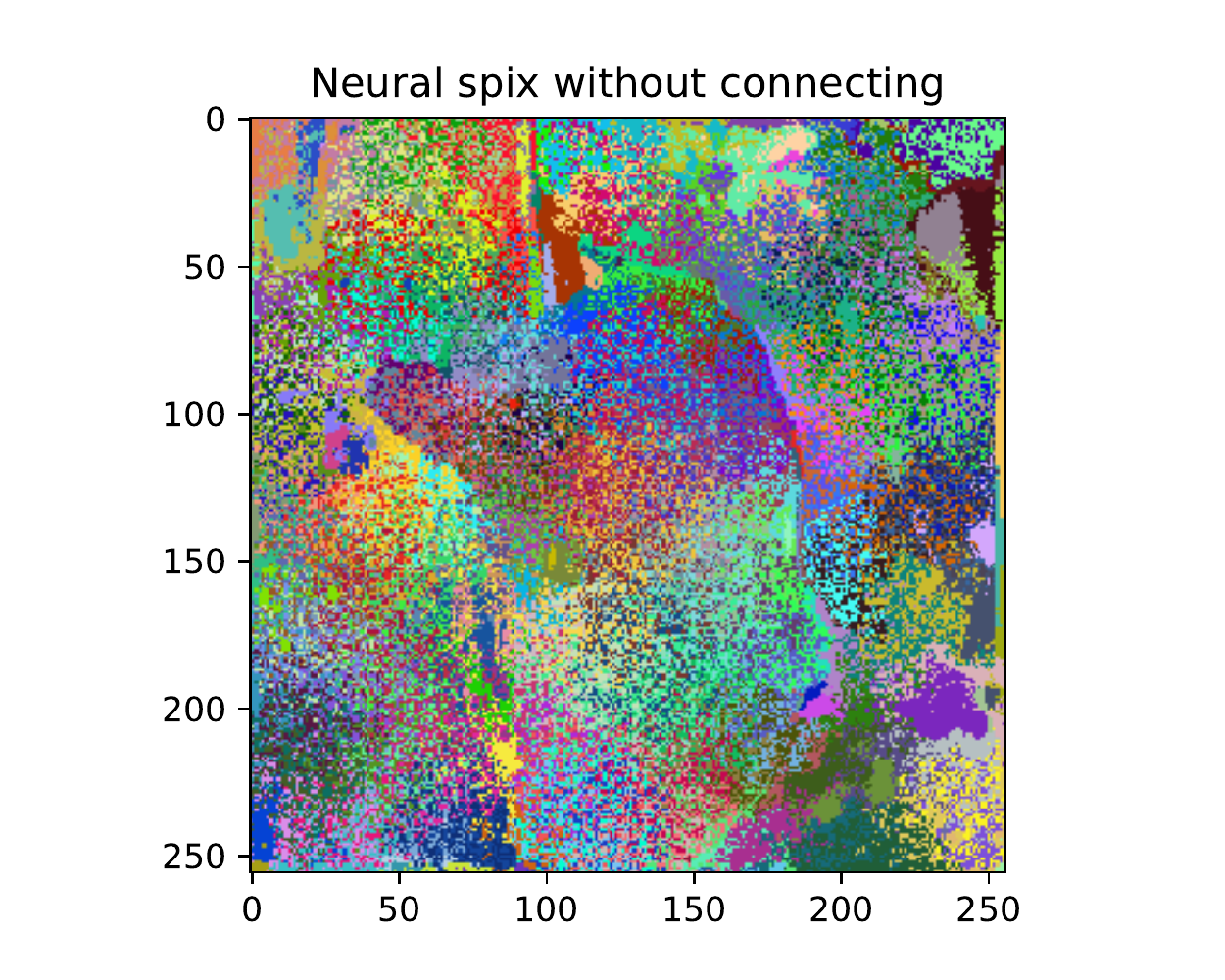}
		\subcaption{At least 2 different distances}
	\end{subfigure}
	\hfill
	\begin{subfigure}[h]{0.32\linewidth}
		\includegraphics[trim={2.72cm 1.42cm 2.72cm 1.42cm},clip,width=\linewidth]{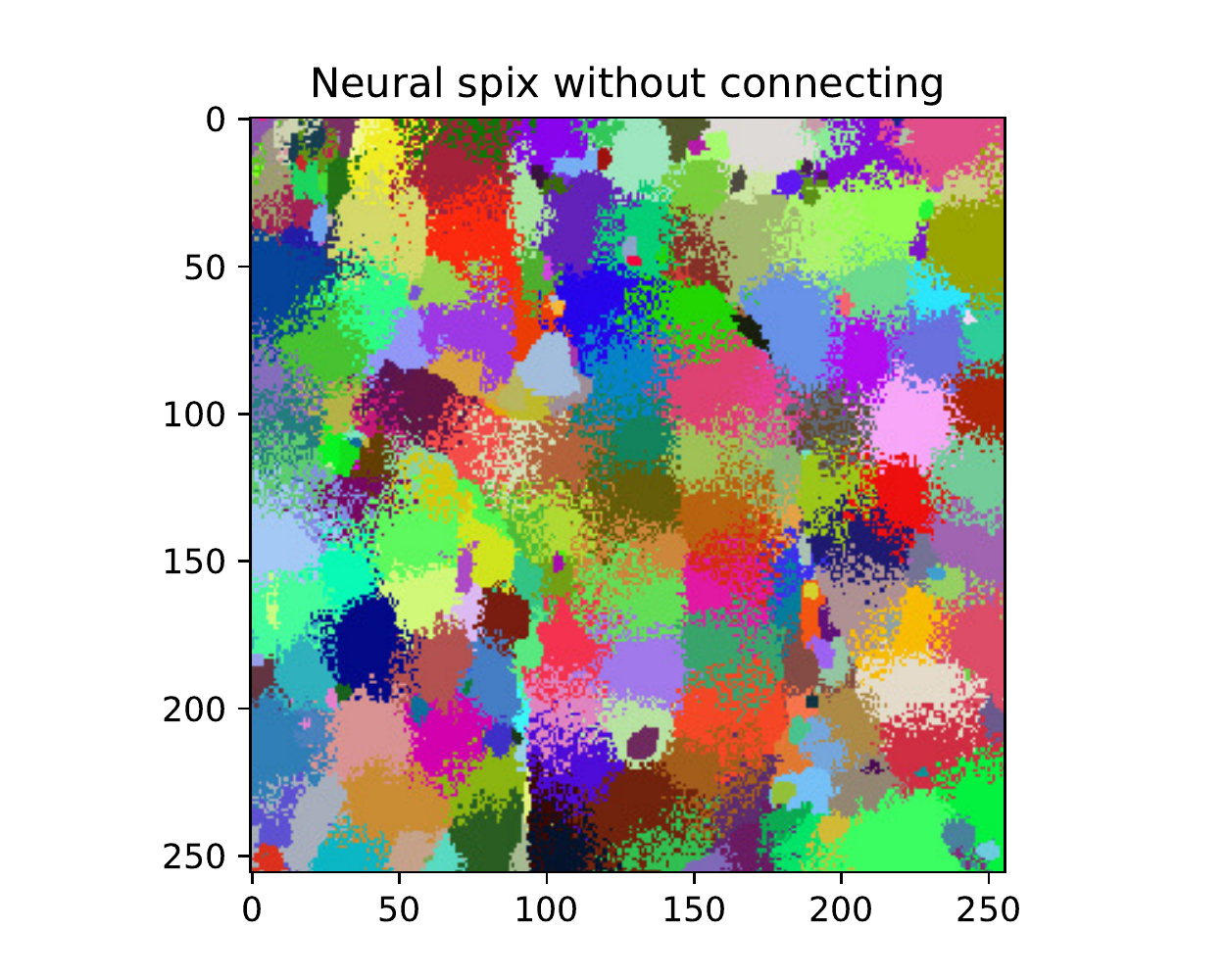}
		\subcaption{At least 4 different distances}
	\end{subfigure}
	\hfill
	\begin{subfigure}[h]{0.32\linewidth}
		\includegraphics[trim={2.72cm 1.42cm 2.72cm 1.42cm},clip,width=\linewidth]{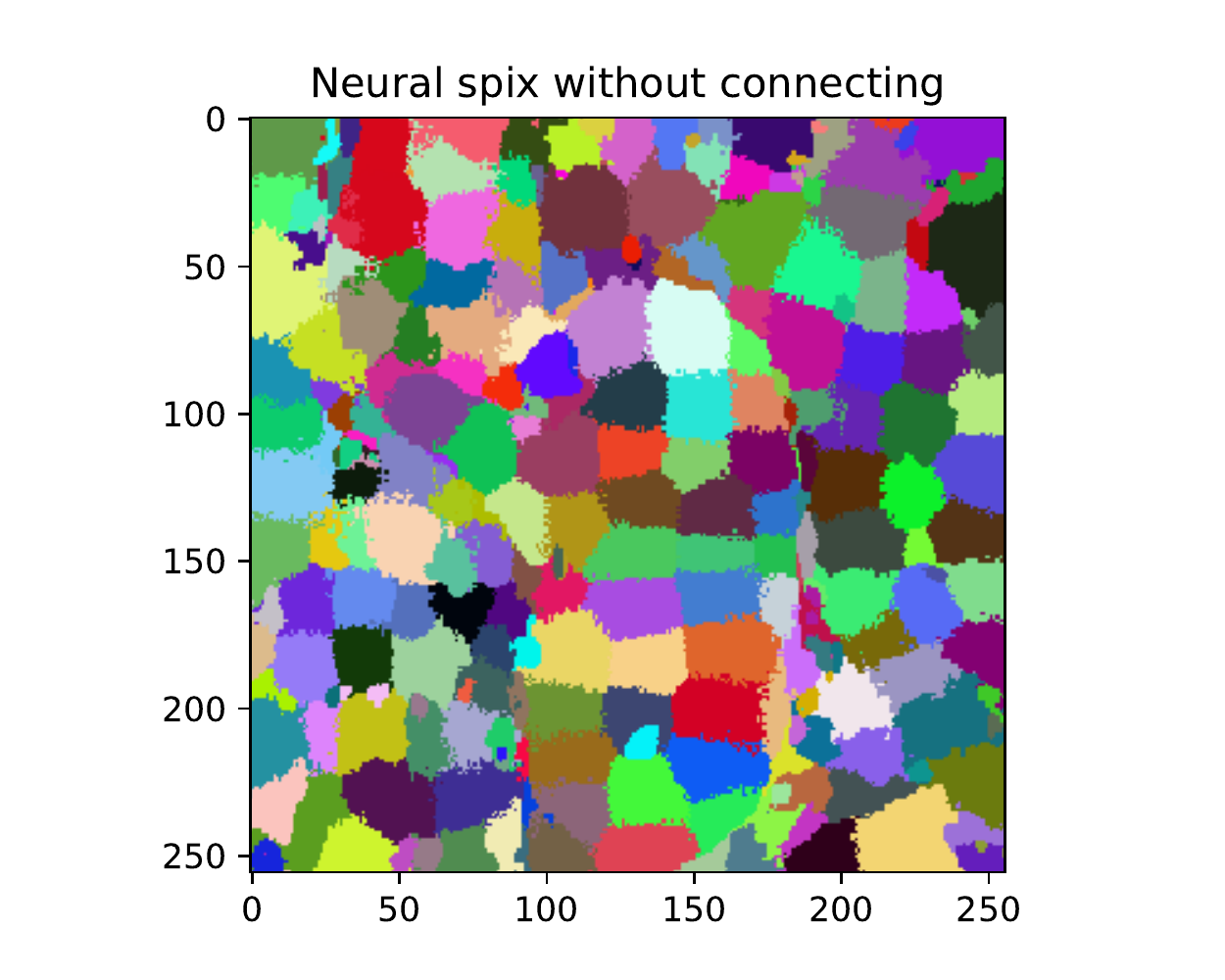}
		\subcaption{At least 6 different distances}
	\end{subfigure}
	\caption{Superpixel output for hard-example mining. Restricting the the label set by removing ambiguous labels produces more compact superpixels.}
	\label{fig:unambiguousoutput}
	\vspace{-0.5cm}
\end{figure}

%% This figure is later in the text, but placed in front so the two hard-mining figures appear on the same page!
\begin{figure}[tb]
	\centering
    \captionsetup{font=small}
	\captionsetup[sub]{font=footnotesize,labelfont={it,it}}
	\begin{subfigure}[h]{0.32\linewidth}
		\includegraphics[trim={3.72cm 1.42cm 3.72cm 1.42cm},clip,width=\linewidth]{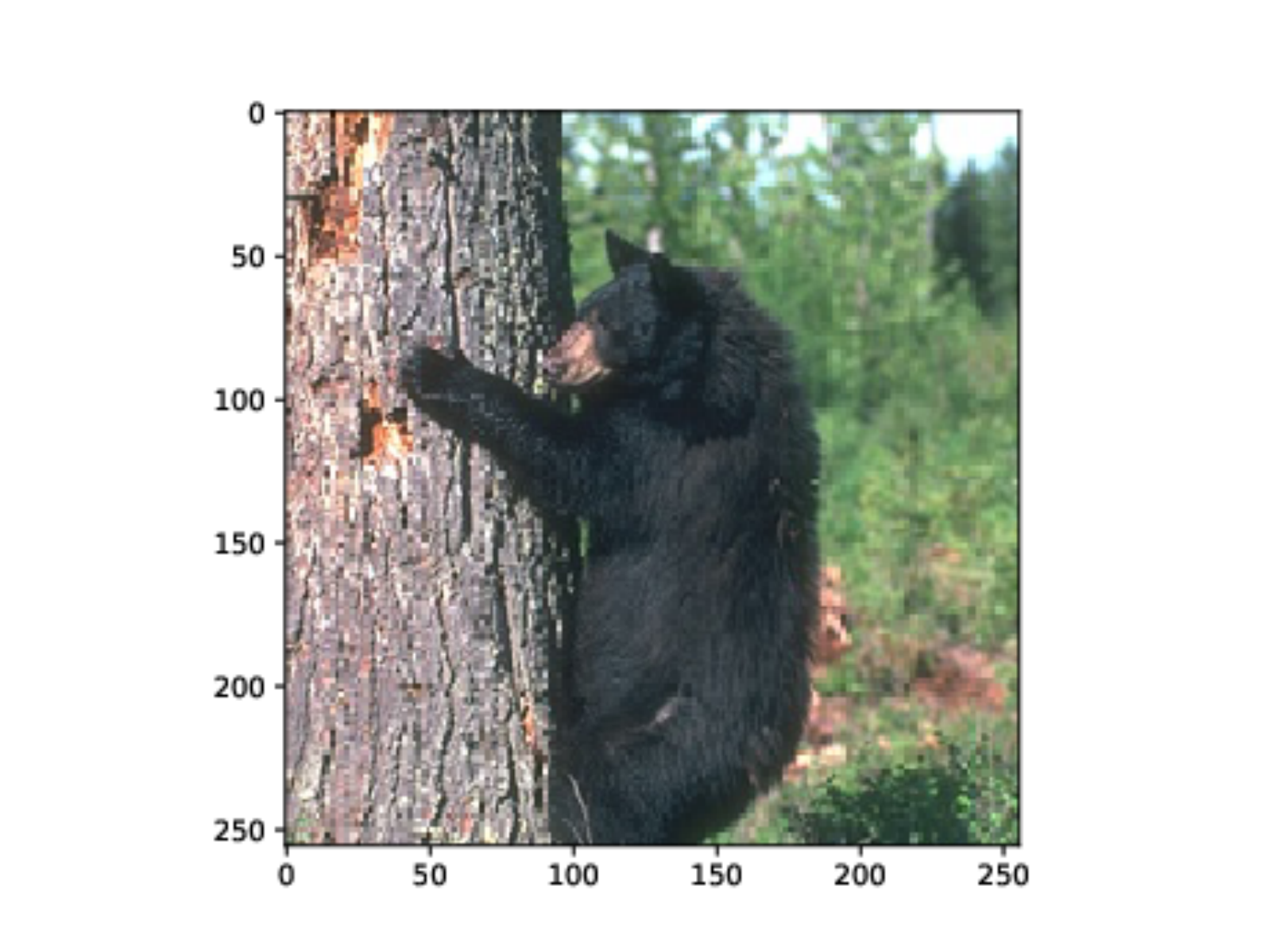}
		\subcaption{Input image}
	\end{subfigure}
	\hfill
	\begin{subfigure}[h]{0.32\linewidth}
		\includegraphics[trim={3.72cm 1.42cm 3.72cm 1.42cm},clip,width=\linewidth]{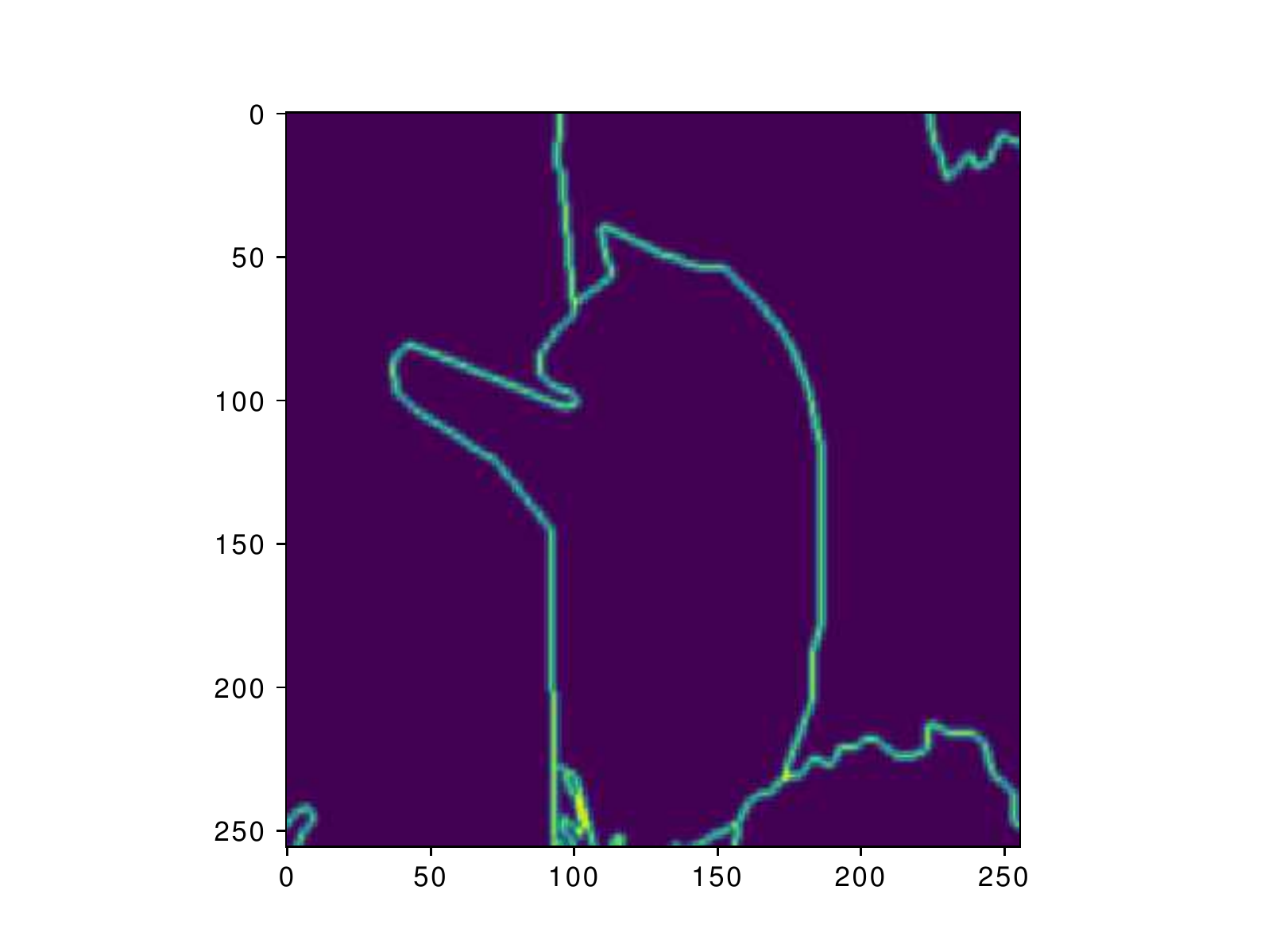}
		\subcaption{Single GT edge map}
	\end{subfigure}
	\hfill
	\begin{subfigure}[h]{0.32\linewidth}
		\includegraphics[trim={3.72cm 1.42cm 3.72cm 1.35cm},clip,width=\linewidth]{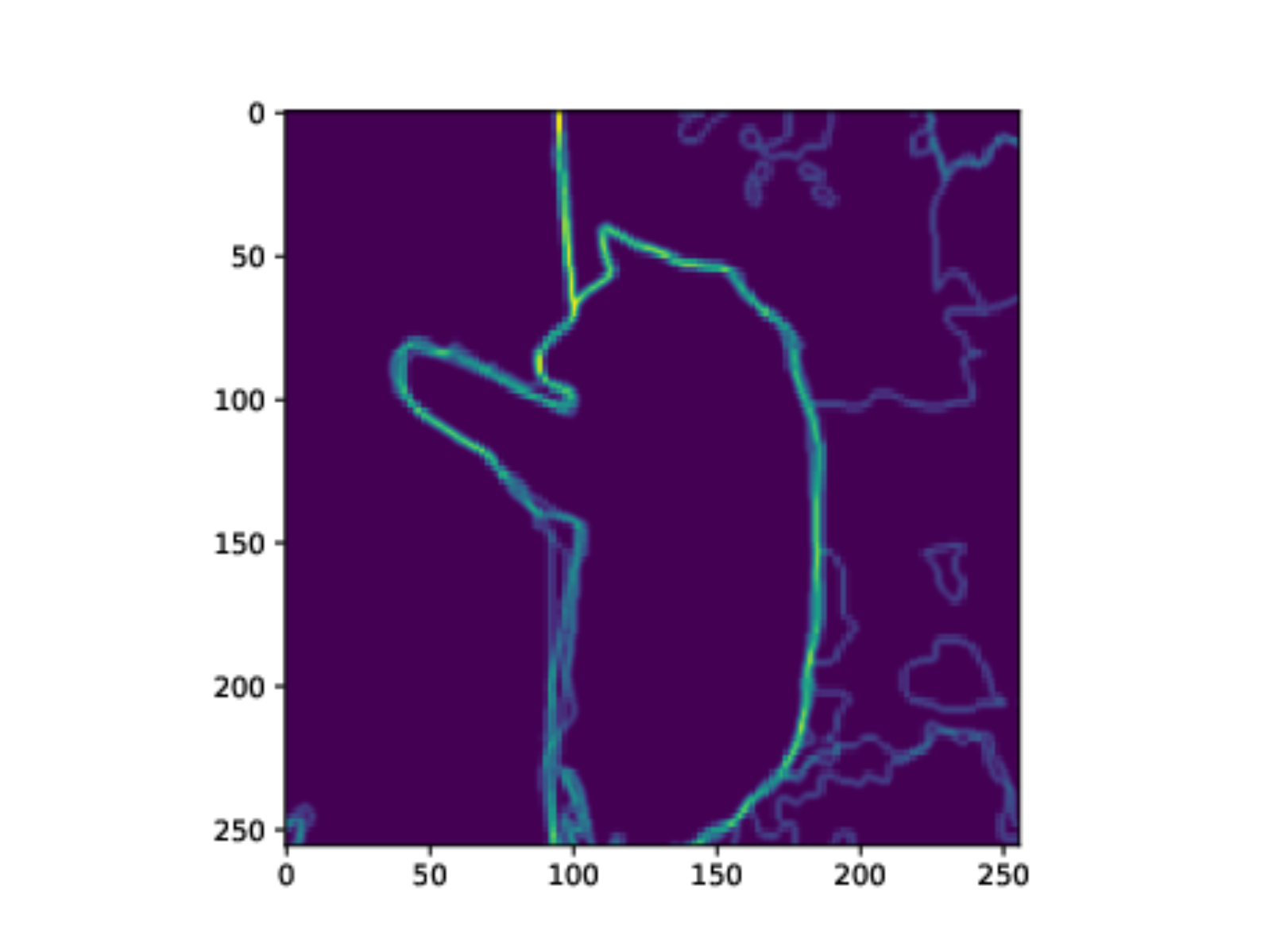}
		\subcaption{Combined edge map}
	\end{subfigure}
	\caption{BSDS500: multiple ground truth annotations are combined in a single edge map. Stronger borders in the image have stronger appearance in the edge map. }
	\label{fig:BSDS500multi}
\end{figure}
\subsection{BSDS ground truth edges}
More refined semantic segmentations provide more accurate labels. We considered several semantic segmentation datasets: PASCAL VOC~\cite{pascal-voc-2012}, Cityscapes~\cite{cityscapes} and BSDS500~\cite{bsds500}. Cityscapes and BSDS both have high-quality ground truth annotations, but BSDS has multiple of them for a single image.  Typically, object borders in natural images are not clearly delineated and multiple independent ground truth segmentations help to handle these cases. We combine the 5 individual ground truth annotations in a single ground truth edge map (\reff{BSDS500multi}). %Delineated areas can be seen as independent ground truth segments. 
This also defines a new distance measure: more edges between a pixel and cluster indicate a greater distance and less likelihood to be assigned to that cluster.

\section{Training a distance measure}

The proposed network interprets the classification task as a typical deep learning problem. We were not able to  replicate the SLIC distance measure \textit{exactly}, although superpixel output was similar. We note that the SLIC distance measure could be perfectly replicated by squaring each element of the input vector and removing the batchnorm layer: the elements of the input vector then become the individual terms of the SLIC distance measure. By making the different parts of the network independent, the trained modules can be seen as distance functions~(\reff{inputvector_asslic}). The network then learns a regression by training a classification. We verified that the network can almost perfectly replicate the SLIC distance measure (Table \ref{table:labeltrainingresultsvalidationloss}). 
When using a single linear layer, the network in fact learns the weights of Equation \ref{formula:slicweighteddsitance}. These weights can then be integrated in the top-down approach of SLIC, resulting in a very efficient trainable superpixel algorithm running on CPU. 

\begin{figure}[tbh]
	\captionsetup{font=small}
	\captionsetup[sub]{font=footnotesize,labelfont={it,it}}
	\centering
	\includegraphics[trim={0cm 0cm 0cm 0cm}, clip, width=1\linewidth]{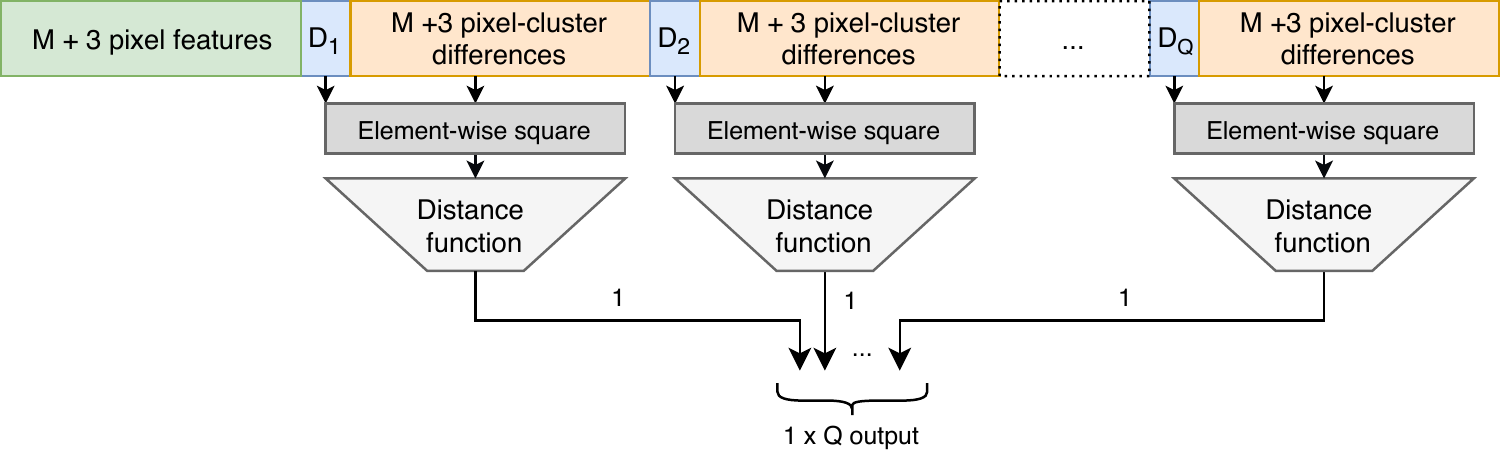}
	\caption{Classifier with distance function modules. The amount of clusters considered during classification can be changed easily. The network learns a distance function between clusters and pixels using classification labels.}
	\label{fig:inputvector_asslic}
\end{figure}

\begin{table*}[b]
	\centering
	\caption{Performance comparison of superpixel methods in this work}
	{\vspace*{-0.2cm}500 BSDS color images, superpixel size 16, compactness 10\vspace*{0.2cm}} 
	\label{table:resultsresults}
	\maxsizebox{\linewidth}{!}{%
	\begin{tabular}{lHllllll}
	    \toprule
		\textbf{Method}                                        & \textbf{IoUu} & \textbf{IoU} & \textbf{Rec} & \textbf{MDE} & \textbf{UE} & \textbf{CO}                \\ \midrule
		\multicolumn{1}{l}{SLIC (reference)}    & 0.906         & 0.907        & 0.809        & 0.911       & 0.101      & \multicolumn{1}{l}{0.324} \\
		\multicolumn{1}{l}{Manual tuning: inclusion of features using an extended distance measure}  & 0.909        & 0.913       & 0.819        & 0.870        & 0.095      & \multicolumn{1}{l}{0.328} \\
		\multicolumn{1}{l}{Deep learning classification network: SLIC with GT-corrected and hard-example labels ($X = 3$)}   & 0.907        & 0.910       & 0.800        & 0.942        & 0.097      & \multicolumn{1}{l}{0.328} \\
		\multicolumn{1}{l}{Deep learning classification network: Weakly supervised hard-example labels ($X = 6$)}  & 0.912        & 0.913       & 0.796        & 0.954         & 0.094        & \multicolumn{1}{l}{0.307 } \\
		\multicolumn{1}{l}{1-layer regression network}   &0.910        &0.912        &0.819        &0.874        &0.095       & \multicolumn{1}{l}{0.325} \\
		\multicolumn{1}{l}{3-layer regression network}     &0.910        &0.912       &0.824        &0.855        &0.094      & \multicolumn{1}{l}{0.320} \\
		\bottomrule
	\end{tabular}
	}
	\vspace{-0.6cm}
\end{table*}

\section{Evaluation and results}
\label{sec:resultslast}

\subsection{Metrics}
%\label{sec:evaluation1}
Superpixel performance is evaluated on 500 BSDS500~\cite{bsds500} color images. Superpixels are evaluated with size $16$, compactness $10$ (determined optimal for the standard SLIC) and 5 clustering iterations.  %and their ground truth segmentations. %Each image contains 5 segmentation ground truth on average, and individual scores are averaged.
We use several metrics common in superpixel evaluation: 
\textit{Boundary recall (Rec)} represents the adherence to ground truth boundaries (higher is better). \textit{Mean distance to edge (MDE) \cite{benesova2014fast}} measures the average distance between the ground truth border and closest superpixel edge (lower is better). Superpixel leakage into different ground truth segments is quantified by the \textit{undersegmentation error (UE)} (lower is better). Multiple variants exist, we use the definition of Neubert and Protzel \cite{neubert2012superpixel}.
The regularity and compactness of superpixels is measured by the \textit{compactness (CO)} metric \cite{compactness}. More regular superpixels are generally preferred. For a fair comparison, the compactness parameters of different methods are chosen so their resulting output compactness is similar. 
We define an additional \textit{intersection-over-union (IoU)} metric similar to the one often used in segmentation benchmarks. This metric measures the maximum achievable performance when using superpixels in a segmentation pipeline. 
% Call the ground truth segmentation $G$. Each superpixel $S_j$ in $S$ gets the label of the most overlapping ground truth segment $G_j$. This defines a segmentation map $T$, used to calculate the IoU score:
% \begin{equation}
% IoU(G, T) = \frac{1}{N}\sum_{G_j}  |G_j| \frac{T_j\cap G_j}{T_j \cup G_j}
% \end{equation}

% The advantage of this metric when evaluating the performance of clustering-based superpixel algorithms is that the IoU metric does not require connected superpixels and avoids the influence of the post-processing algorithm. When evaluating on unconnected clusters, we call the metric \textit{unconnected intersection-over-union (IoUu)}.

\subsection{Extended distance measure with manual tuning}
\label{sec:manualevaluation}
As a first experiment, we evaluate the inclusion of scattering features in the extended distance measure for SLIC (Section~\ref{sec:simpleintegration}). The scattering transformation is applied on the lightness channel $L$ of the image (converted to the $CIELAB$ color space) and we manually select the most important representations.  We refer to this method as \textit{\lq Manual tuning\rq} and Table~\ref{table:resultsresults} shows that all metrics are improved. Mainly the mean distance to edge and undersegmentation metrics are impacted: the low-resolution features do not help at a pixel-scale level, but avoid superpixel leakage. The difference is larger at lower compactness values (\reff{comp_comp}). Evaluating the methods for their own optimal compactness, improvement of MDE is 9.4\% compared the 4.3\% improvement for $\sigma = 10$. The approach with scattering features benefits from the increased flexibility, while SLIC performance decreases. Superpixels incorporating deep representations also consistently perform better (\reff{comp_consis}): most images are slightly improved. % the score is never significantly decreased. 
In addition, we experimented with greyscale images and the effect of scattering features is even stronger. 
%For $\sigma = 10$, the MDE performance improvement is 4.3\%, while for the very low compactness setting of $\sigma = 1$ this is 17.7\%. 
 
% \begin{table}[tb]
% 	\centering
% 	\caption{Superpixel evaluation with scattering features}
% 	{\vspace{-0.2cm}500 BSDS images, superpixel size 16, compactness 10, 5 iterations\vspace{0.2cm}} 
% 	\label{table:colorbsds}
% \begin{tabular}{@{}lHccccc@{}}
% \toprule
%             & IoUu  & IoU   & Rec   & MDE   & UE     & CO    \\ \midrule
% SLIC        & 0.906 & 0.907 & 0.809 & 0.911 & 0.1005 & 0.324 \\
% SLIC + features & 0.909 & 0.913 & 0.819 & 0.870 & 0.0945 & 0.328 \\ \bottomrule
% \end{tabular}
% \end{table}

\begin{figure}[tbh]
	\captionsetup{font=small}
	\captionsetup[sub]{font=footnotesize,labelfont={it,it}}
	\centering
% 	\begin{subfigure}[h]{0.45\linewidth}
% 		\includegraphics[trim={0.72cm 0.42cm 0.72cm 0.42cm}, clip,width=\linewidth]{comparison_2_plots/comp_iou.eps}
% 	\end{subfigure}
	%\hfill
	\begin{subfigure}[h]{0.49\linewidth}
		\includegraphics[trim={0.1cm 0.05cm 2.5cm 0.2cm}, clip,width=\linewidth]{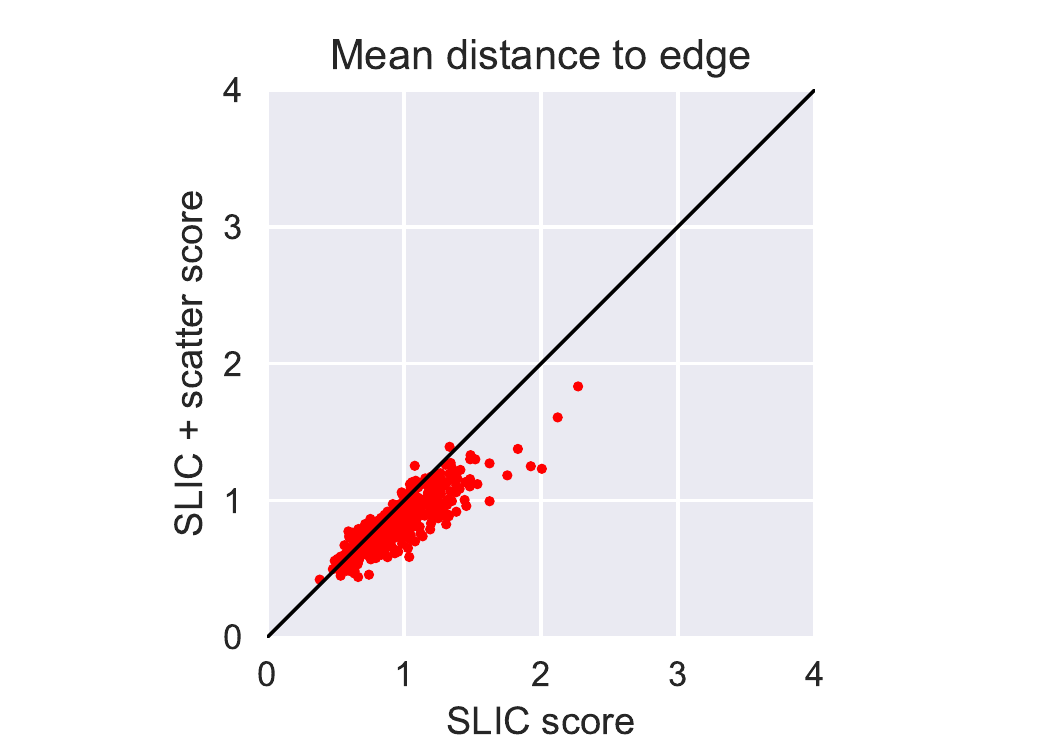}
	\end{subfigure}
	\begin{subfigure}[h]{0.49\linewidth}
		\includegraphics[trim={0.2cm 0.05cm 2.5cm 0.2cm}, clip,width=\linewidth]{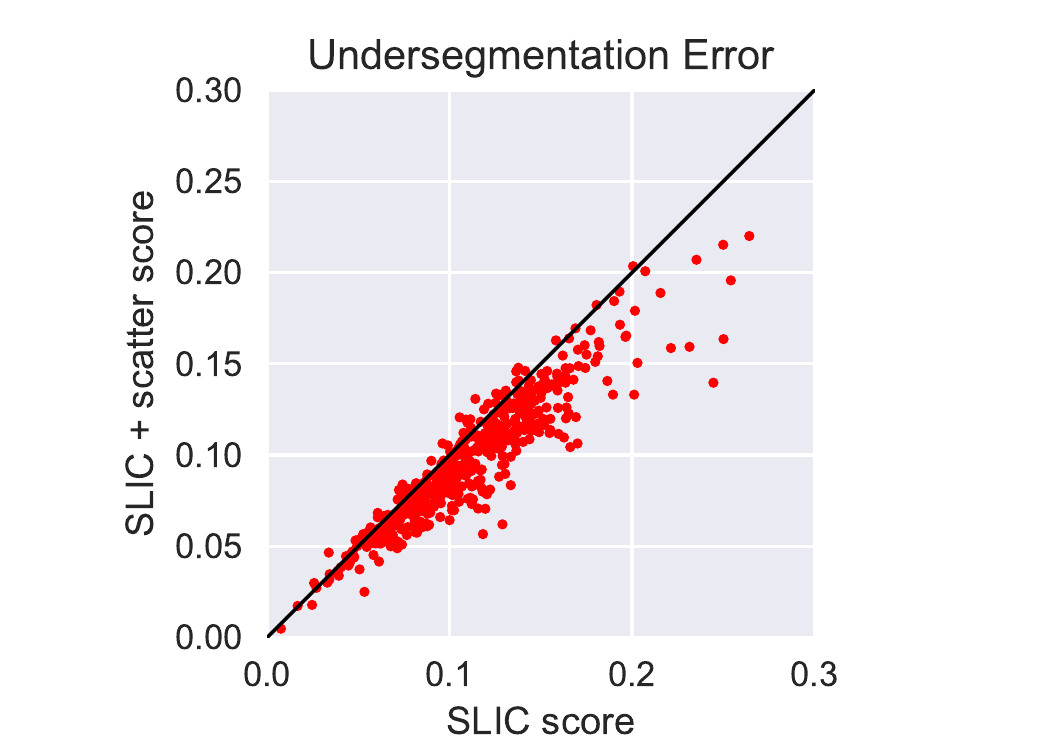}
	\end{subfigure}
    \vspace*{-0.1cm}
	\caption{Consistency of results: scores per image for the mean distance to edge (lower is better) and undersegmentation error (lower is better) metric. The SLIC implementation with scattering features and manually tuned weights scores consistently better.  }
	\label{fig:comp_consis}
\end{figure}
\begin{figure}[tbh]
	\captionsetup{font=small}
	\captionsetup[sub]{font=footnotesize,labelfont={it,it}}
	\centering
% 	\begin{subfigure}[h]{0.45\linewidth}
% 		\includegraphics[trim={0.72cm 0.42cm 0.72cm 0.42cm}, clip,width=\linewidth]{comparison_2_plots/comp_ioum.eps}
% 	\end{subfigure}
	%\hfill
	\begin{subfigure}[h]{0.49\linewidth}
		\includegraphics[trim={0.00cm 0.02cm 1.05cm 0.2cm}, clip,width=\linewidth]{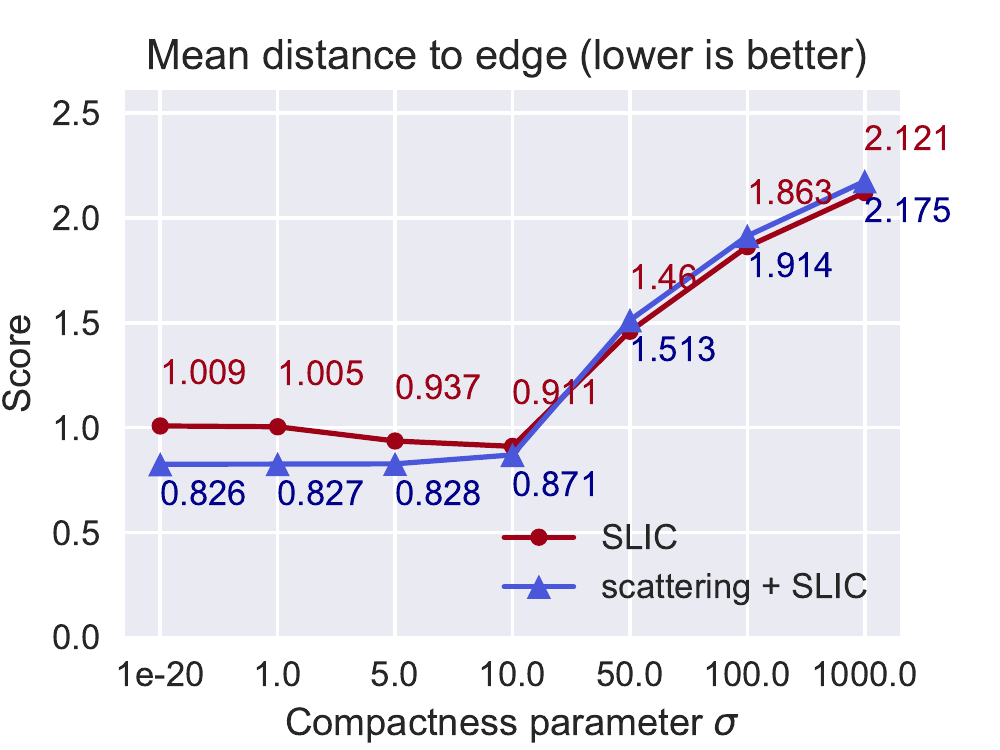}
	\end{subfigure}
	\hfill
	\begin{subfigure}[h]{0.49\linewidth}
		\includegraphics[trim={0.00cm 0.02cm 1.05cm 0.2cm}, clip,width=\linewidth]{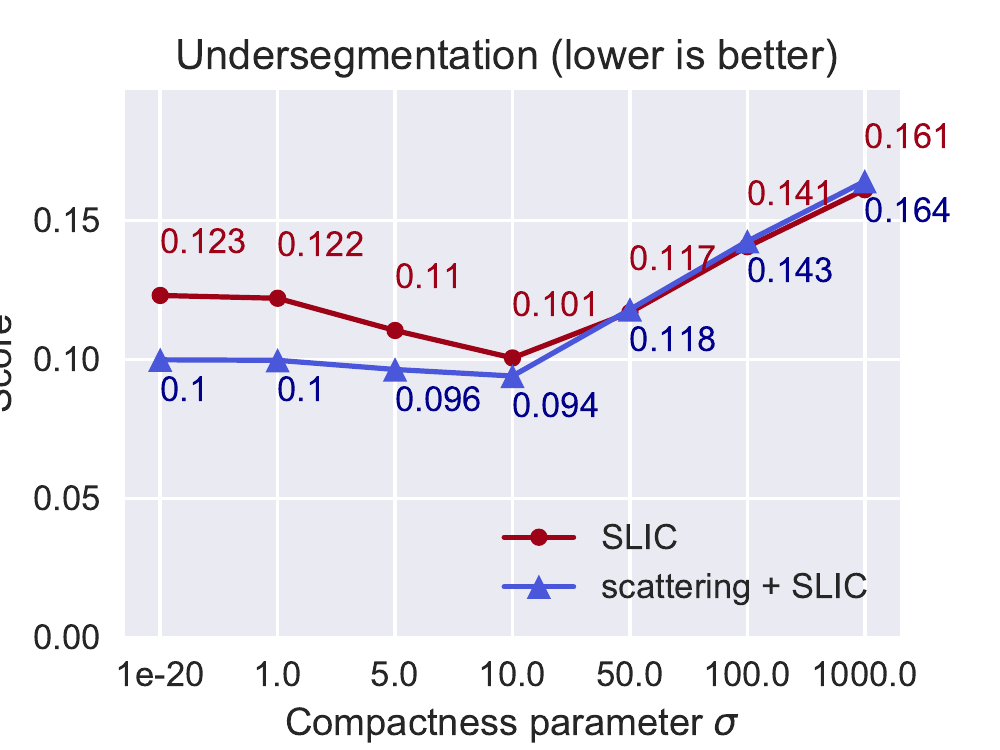}
	\end{subfigure}
    \vspace*{-0.1cm}
	\caption{Influence of compactness parameter $\sigma$ on MDE and UE metrics for the manually tuned method.}
	\label{fig:comp_comp}
	\vspace*{-0.5cm}
\end{figure}

\subsection{Trainable superpixels}
Trainable superpixels should be able to improve superpixel quality without having to manually tune the distance measure weights.
Quality assessment of the trainable superpixels is a three-stage process: a label set is generated, a classifier is trained on these labels and the superpixel algorithm using the trained classifier is evaluated.  We selected the most promising label methods for evaluation on $256 \times 256$ BSDS500 images and tested scattering and ENet features. The 243 scattering features have a receptive field of $4 \times 4$ and spatial dimensions of $64 \times 64$. The ENet features are extracted from the first convolutional layer, designed to be feature extractor and consisting of filters having a receptive field of $3 \times 3$. They have a better spatial resolution of size $128 \times 128$, but there are only 16 features. 

We selected a simple network with an architecture as in \reff{inputvector_asdr}, where the dimensionality reducers \textit{DRP/DRC} are 2-layer networks (hidden layers of 100 and 15 neurons) and the classification network \textit{FC} is a 4-layer network (hidden layers of size 120, 105 and 15, output layer of 7 neurons). All activation functions are rectified linear units (ReLU). We call this network the \textit{\lq Deep learning classification network\rq}.
We also test regression architectures as in \reff{inputvector_asslic} with a single linear layer for the distance measure module: this is in fact just a weighted addition of the squared pixel-cluster differences. 
%For evaluation, trained weights are integrated in top-down SLIC. 
This approach is called \textit{\lq 1-layer network\rq}. In addition, we evaluate a network where the single layer is expanded to 3 layers \textit{(\lq 3-layer\rq)}.

We trained on several label methods and experimented with different variations of hard-example mining for both SLIC-based labels and weakly supervised labels. Our experiments found that more engineered methods performed better. The best SLIC-based method corrects labeling mistakes with the segmentation ground truth and applies hard-example mining, with parameter $X = 3$, in order to remove ambiguous labels.
The best weakly supervised label method also removes ambiguous labels, but with parameter $X = 6$, retaining clusters lying in at least 6 different ground truth segments. 

\subsection{Validation loss}
\begin{table}[tb]
	\captionsetup{font=small}
	\captionsetup[sub]{font=footnotesize,labelfont={it,it}}
	\centering
	\caption{Validation loss for different training methods, features and networks}
    
\label{table:labeltrainingresultsvalidationloss}

\maxsizebox{\columnwidth}{!}{%
\begin{tabular}{@{}lcHc@{}}
\toprule
Label method                                           & Scat F & SFsq  & ENet F \\ \midrule
\multicolumn{4}{c}{\textbf{Deep learning classification-based network}} \\
SLIC replication                                       & 0.184 & 0.193 & 0.210  \\
%SLIC with GT-corrected labels     & 0.275 & 0.348 & 0.252  \\%
SLIC with GT-corrected hard-example labels ($X = 3$) & 0.786 & 0.786 & 0.746  \\
%SLIC with GT-based hard labels                        & 1.669 & 1.819 & 1.552  \\
%Weakly: BSDS-edges + unambiguous labels ($X = 4$)      & 1.345 & 1.524 & 1.530  \\
Weakly supervised hard-example labels ($X = 6$)      & 1.413 & 1.428 & 1.400  \\ \midrule

\multicolumn{4}{c}{{\vspace*{0.2cm}\textbf{1-layer regression-based network}\vspace*{0.00cm}}} \\
SLIC replication  & 0.016    &                    & 0.035         \\
SLIC with GT-corrected hard-example labels ($X = 3$)  & 0.970     &                   & 2.885         \\ \midrule

\multicolumn{4}{c}{\vspace*{0.2cm}\textbf{3-layer regression-based network}\vspace*{0.0cm}} \\
SLIC with GT-corrected hard-example labels ($X = 3$)  & 0.731         &               & 0.960         \\ \bottomrule

\end{tabular}
}
\vspace{-0.6cm}
\end{table}
%Comparing the validation loss between different labeling methods is meaningless: harder methods will have a higher validation loss. 

As different labeling methods employ different loss functions, we cannot directly compare the values of these loss functions on the validation set. For a single label method, a comparison between network architectures and features is possible and serves as an indication for resulting superpixel quality. Table~\ref{table:labeltrainingresultsvalidationloss} shows that scattering features and ENet features achieve similar validation losses in most cases. Unsurprisingly, the 3-layer regression network performs better than the 1-layer one, and it also performs slightly better than the classification-based network that used batch normalization and dimensionality reduction modules.

\subsection{Superpixel quality}
\begin{figure}[tbh]
	\captionsetup{font=small}
	\captionsetup[sub]{font=footnotesize,labelfont={it,it}}
	\centering
% 	\begin{subfigure}[h]{0.45\linewidth}
% 		\includegraphics[trim={0.72cm 0.42cm 0.72cm 0.42cm}, clip,width=\linewidth]{comparison_2_plots/comp_ioum.eps}
% 	\end{subfigure}
	%\hfill
	\begin{subfigure}[h]{0.49\linewidth}
		\includegraphics[trim={1.00cm 0.02cm 1.05cm 0.2cm}, clip,width=\linewidth]{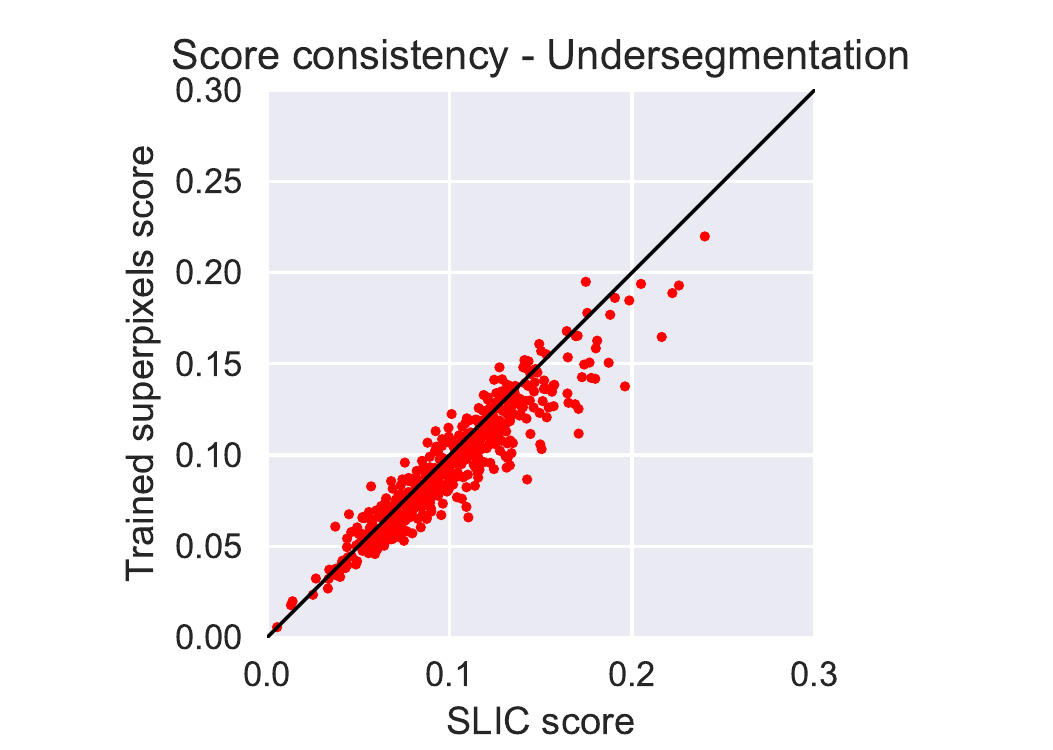}
	\end{subfigure}
	\hfill
	\begin{subfigure}[h]{0.49\linewidth}
		\includegraphics[trim={1.00cm 0.02cm 1.05cm 0.2cm}, clip,width=\linewidth]{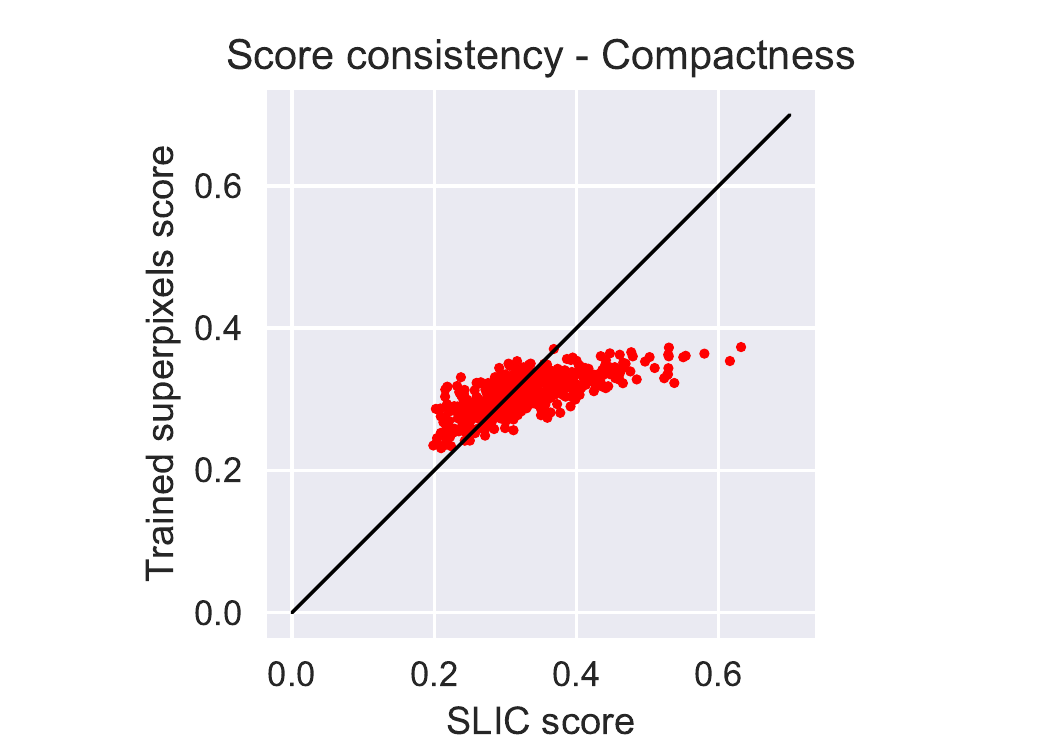}
	\end{subfigure}
    \vspace*{-0.1cm}
	\caption{Scatter plot of the scores per image for undersegmentation (lower is better) and compactness metrics. The trainable approach using weakly supervised labels improves the undersegmentation error quite consistently and the produced superpixels have a low variation in compactness compared to SLIC.}
	\label{fig:trainablescoresweakly}
	%\vspace*{-0.5cm}
\end{figure}

The superpixel quality for each of these methods is compared in Table~\ref{table:resultsresults}. The 3-layer regression-based network, having the lowest validation loss, also achieves the best metric scores. Superpixel quality is improved over standard SLIC and also over the manually tuned method of Section~\ref{sec:simpleintegration}. Comparing methods visually (\reff{spixoutput55}) shows that the manually tuned method tends to concentrate superpixels around object borders. This effect is not seen in the trained superpixels. During evaluation of manually tuned superpixels, we already discovered that the extra features mainly influence the mean distance to edge and undersegmentation metrics and the same effect can be seen here.
%The regression approach, where a distance function is learned, seems optimal for this problem. 
 The weakly supervised method has surprisingly similar scores to SLIC and during our tests we noticed that the variation in compactness was much lower (Fig \ref{fig:trainablescoresweakly}). 
 
%In addition, we noticed that the weakly supervised approach generated superpixels with a smaller compactness spread than SLIC.

\section{Conclusion}
Superpixels are image priors that tend to transfer across tasks. This works elaborates on a trainable approach for superpixels incorporating deep image representations. We introduce several new ideas not yet addressed in research: we include deep representations in a superpixel algorithm, build a set of superpixel training labels from segmentation annotations and devise a trainable superpixel algorithm. We demonstrate that a simple inclusion of deep representations by extending the SLIC distance measure improves superpixel quality in a consistent way. The trainable approach can surpass the scores of the simple inclusion, but requires appropriate training labels.  
The performance increase could be limited by the dataset and features used in our experiments. We used natural images, which have a high variability in features. We believe larger performance increases can be achieved by targeting specific modality, such as medical imaging. More specialized features can be incorporated, possibly having a less restricted receptive field than the scattering features. We hope that our analysis paves the way to the inclusion of trainable superpixels in deep learning pipelines.

\begin{figure}[tb]
    \captionsetup{font=small}
	\captionsetup[sub]{font=footnotesize,labelfont={it,it}}
	\captionsetup[subfigure]{labelformat=empty}
	\centering
	\begin{subfigure}[h]{0.325\linewidth}
		\includegraphics[trim={4.2cm 3cm 4.8cm 2cm}, clip,width=\linewidth]{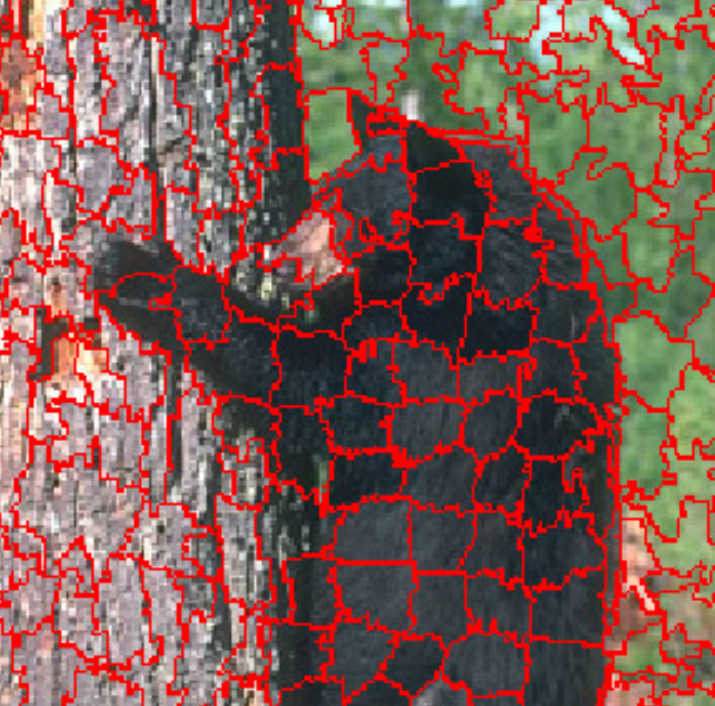}
		\subcaption{SLIC}
		\label{fig:spixoutput55a}
	\end{subfigure}
	\begin{subfigure}[h]{0.325\linewidth}
		\includegraphics[trim={4.2cm 3cm 4.8cm 2cm}, clip,width=\linewidth]{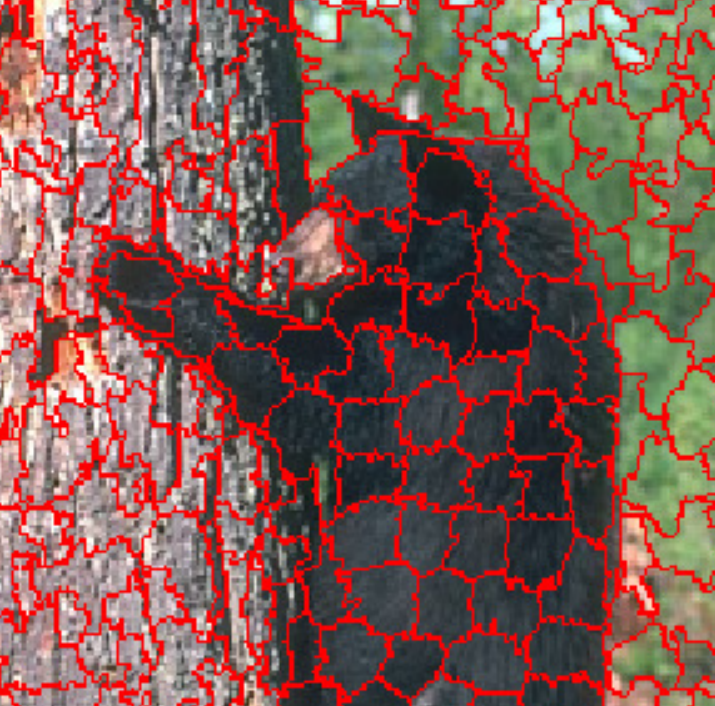}
		\subcaption{Manual tuning}
		\label{fig:spixoutput55c}
	\end{subfigure}
	\begin{subfigure}[h]{0.325\linewidth}
		\includegraphics[trim={4.2cm 3cm 4.8cm 2cm}, clip,width=\linewidth]{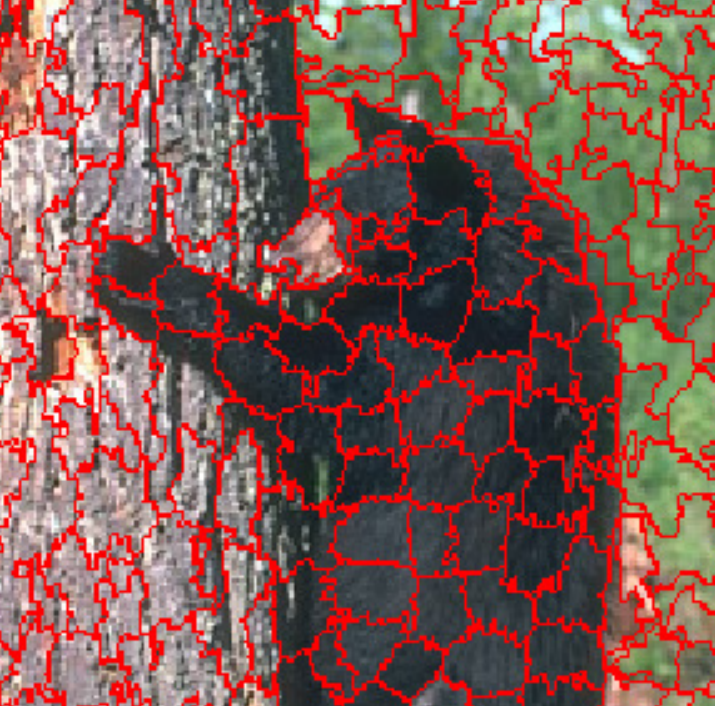}
		\subcaption{Trainable (3-layer)}
		\label{fig:spixoutput55b}
	\end{subfigure}
	\caption{Outputs of different superpixel methods. The manual tuning method tends to concentrate superpixels around object edges, for example at the bear's paw. }%This effect does not happen with trainable superpixels. }
	\vspace{-0.5cm}
	\label{fig:spixoutput55}
\end{figure}

% {\small
% \bibliographystyle{ieee}
% \bibliography{egbib}
% }
{
\small
\printbibliography
}

%\printbibliography

\end{document}